\documentclass{article}

\usepackage{microtype}
\usepackage{graphicx}
\usepackage{booktabs} % for professional tables

\RequirePackage{algorithm}
\RequirePackage{algorithmic}

\usepackage[accepted]{icml2026}

\usepackage{url}
\usepackage{booktabs}       % professional-quality tables
\usepackage{amsfonts}       % blackboard math symbols
\usepackage{microtype}      % microtypography

\usepackage{amsmath}
\usepackage{amssymb}
\usepackage{mathtools}
\usepackage{amsthm}

\usepackage{subfigure}
\usepackage{tikz}
\usepackage{accsupp}
\usepackage{listofitems}
\usepackage{xstring}
\usepackage{varwidth}
\usepackage{stfloats}
\usepackage{wrapfig}
\usepackage{tabularx}
\usepackage{adjustbox}
\usepackage{etoolbox}
\usepackage{hyperref}

\usepackage[capitalize,noabbrev]{cleveref}

\usetikzlibrary{calc,positioning,arrows.meta}

\theoremstyle{plain}

\theoremstyle{definition}

\theoremstyle{remark}

\def \sfscale {0.05}
\newcommand{\snowflake}{%
\begin{tikzpicture}[y=1cm, x=1cm, yscale=\sfscale,xscale=\sfscale, every node/.append style={scale=\sfscale}, inner sep=0pt, outer sep=0pt]
  \path[fill,line cap=,miter limit=10.0] (2.8734, -7.2284) -- (3.7015, -6.7495) -- (3.8391, -5.7362) -- (4.0852, -5.7679) -- (3.9793, -6.5881) -- (4.8366, -6.0934) -- (4.99, -6.3632) -- (4.1301, -6.858) -- (4.908, -7.1834) -- (4.8075, -7.403) -- (3.855, -7.0168) -- (3.0242, -7.4983) -- (3.0242, -7.62) -- (3.855, -8.0989) -- (4.8075, -7.6941) -- (4.908, -7.9322) -- (4.1275, -8.255) -- (4.9874, -8.7498) -- (4.8339, -9.0144) -- (3.974, -8.5196) -- (4.0825, -9.353) -- (3.8365, -9.3821) -- (3.6989, -8.3582) -- (2.8707, -7.8793) -- (2.7702, -7.9507) -- (2.7702, -8.9032) -- (3.5878, -9.5276) -- (3.4343, -9.7261) -- (2.7702, -9.2234) -- (2.7702, -10.2129) -- (2.4712, -10.2129) -- (2.4712, -9.2234) -- (1.7965, -9.7261) -- (1.6457, -9.5276) -- (2.4712, -8.9032) -- (2.4712, -7.9507) -- (2.3627, -7.8793) -- (1.5319, -8.3582) -- (1.4023, -9.3821) -- (1.1509, -9.353) -- (1.2568, -8.5196) -- (0.3969, -9.0117) -- (0.2487, -8.7551) -- (1.1086, -8.2524) -- (0.3387, -7.9349) -- (0.4392, -7.702) -- (1.3838, -8.0989) -- (2.2093, -7.62) -- (2.2093, -7.4983) -- (1.3811, -7.0168) -- (0.4366, -7.4136) -- (0.336, -7.1834) -- (1.1033, -6.858) -- (0.2461, -6.3632) -- (0.3942, -6.0934) -- (1.2568, -6.5881) -- (1.1509, -5.7679) -- (1.4023, -5.7362) -- (1.5346, -6.7495) -- (2.3627, -7.2284) -- (2.4686, -7.1702) -- (2.4686, -6.2177) -- (1.6457, -5.5933) -- (1.7965, -5.3949) -- (2.4712, -5.8976) -- (2.4712, -4.908) -- (2.7622, -4.908) -- (2.7622, -5.8976) -- (3.4369, -5.3949) -- (3.5904, -5.5933) -- (2.7649, -6.2177) -- (2.7649, -7.1702) -- cycle;
\end{tikzpicture}%
\xspace%
}

\DeclareUnicodeCharacter{2744}{\rlap{\textcolor{white}{F}}\snowflake}

\usepackage{hyperref}

\usepackage{etoolbox}

\usepackage{savesym,multirow,url,xspace,graphicx,hyperref,enumitem,color}
\usepackage[titletoc]{appendix}
\usepackage{subcaption}
\usepackage{dsfont}
\usepackage{multicol}

\usepackage{tikz}
\usetikzlibrary{automata, positioning, arrows}
\tikzset{
	->, % makes the edges directed
	every state/.style={thick, fill=gray!10}, % sets the properties for each ’state’ node
	initial text=$ $, % sets the text that appears on the start arrow
}
\usetikzlibrary{arrows.meta}
\usepackage{pgfplots}
\pgfplotsset{width=10cm,compat=1.9}
\usepackage{diagbox}

\usepackage{amsmath,amssymb,xfrac,amsthm}
\usepackage{mathtools}
\makeatletter

\newcommand{\Rmnum}[1]{\expandafter\@slowromancap\romannumeral #1@}
\makeatother

\usepackage{enumitem}
\usepackage{wasysym}

\DeclareMathOperator*{\argmin}{arg\,min}
\DeclareMathOperator*{\argmax}{arg\,max}

\newcommand{\Acal}{\mathcal{A}}

\newcommand{\Dcal}{\mathcal{D}}

\newcommand{\Mcal}{\mathcal{M}}
\newcommand{\Ncal}{\mathcal{N}}

\newcommand{\Scal}{\mathcal{S}}
\newcommand{\Tcal}{\mathcal{T}}

\newcommand{\rbar}{\bar{r}}
\newcommand{\rtilde}{\tilde{r}}
\newcommand{\rdagger}{r^\dagger}
\newcommand{\pidagger}{\pi^\dagger}
\newcommand{\Dbar}{\bar{D}}
\newcommand{\Cbar}{\bar{C}}

\newcommand{\E}{\mathbb{E}}
\newcommand{\V}{\text{Var}}
\newcommand{\R}{\mathbb{R}}
\newcommand{\N}{\mathbb{N}}

\usepackage{bm}
\usepackage{bbm}

\newcommand{\norm}[1]{\left\lVert#1\right\rVert}

\newcommand{\INDSTATE}{\STATE\ \ \ \ }

\definecolor{mydarkblue}{rgb}{0,0.08,0.45}
\hypersetup{ %
    pdftitle={Robust In-Context Reinforcement Learning Under Reward Poisoning Attacks},
    pdfsubject={},
    pdfkeywords={in-context reinforcement learning,reward poisoning attacks,meta-learning},
    pdfborder=0 0 0,
    pdfpagemode=UseNone,
    colorlinks=true,
    linkcolor=mydarkblue,
    citecolor=mydarkblue,
    filecolor=mydarkblue,
    urlcolor=mydarkblue,
}

\newcommand{\papertitle}{Robust In-Context Reinforcement Learning Under Reward Poisoning Attacks}

\icmltitlerunning{\papertitle}

\begin{document}

\addtocontents{toc}{\protect\setcounter{tocdepth}{-1}} % This excludes main text from table of contents

\twocolumn[
  \icmltitle{\papertitle}

  % You can specify symbols, otherwise they are numbered in order. Ideally, you
  % should not use this facility. Affiliations will be numbered in order of
  % appearance and this is the preferred way.
  \icmlsetsymbol{equal}{*}
  \icmlsetsymbol{intern}{\textsuperscript{$\dagger$}}

  \begin{icmlauthorlist}
    \icmlauthor{Paulius Sasnauskas}{uoa,amii,intern}
    \icmlauthor{Yiğit Yalın}{mpi}
    \icmlauthor{Goran Radanović}{mpi}
  \end{icmlauthorlist}

  \icmlaffiliation{uoa}{Department of Computing Science, University of Alberta, Edmonton, Canada.}
  \icmlaffiliation{amii}{Alberta Machine Intelligence Institute (Amii), Edmonton, Canada.}
  \icmlaffiliation{mpi}{\mbox{MPI-SWS}, Saarbrücken, Germany}

  \icmlcorrespondingauthor{Paulius Sasnauskas}{sasnausk@ualberta.ca}

  \icmlkeywords{in-context reinforcement learning,reward poisoning attacks,meta-learning}

  \vskip 0.3in
]

% this must go after the closing bracket ] following \twocolumn[ ...

% This command actually creates the footnote in the first column listing the
% affiliations and the copyright notice. The command takes one argument, which
% is text to display at the start of the footnote. The \icmlEqualContribution
% command is standard text for equal contribution. Remove it (just {}) if you
% do not need this facility.

% DO NOT remove the command.
% If you have no special notice, KEEP empty braces:
\printAffiliationsAndNotice{\textsuperscript{$\dagger$}This work was done as a part of an internship project at \mbox{MPI-SWS}.}
% Or, if applicable, use the standard equal contribution text:
% \printAffiliationsAndNotice{\icmlEqualContribution}

\begin{abstract}

We study the corruption-robustness of in-context reinforcement learning (ICRL), focusing on the Decision-Pretrained Transformer \citep[DPT,][]{lee2023dpt}.
To address the challenge of reward poisoning attacks targeting the DPT, we propose a novel adversarial training framework, called Adversarially Trained DPT (AT-DPT). % Decision-Pretrained Transformer
% Our method alternates between optimizing an attacker, which seeks to minimize the true reward of the DPT by poisoning environment rewards, and training the DPT to infer optimal actions from the poisoned data.
Our method simultaneously trains a population of attackers to minimize the true reward of the DPT by poisoning environment rewards, and a DPT model to infer optimal actions from the poisoned data.
% TODO: adaptive-attacker?
% TODO: strong vs. weak? maybe not
% TODO: obtain an algorithm that learns to learn against attacks
We evaluate the effectiveness of our approach against standard bandit algorithms, including robust baselines designed to handle reward contamination.
Our results show that AT-DPT significantly outperforms them in bandit settings under a learned attacker, and generalizes to more complex environments such as adaptive attackers and MDPs.
% We additionally evaluate AT-DPT on an adaptive attacker, and observe similar results.
% Furthermore, we extend our evaluation to MDPs, confirming that the robustness observed in bandit scenarios generalizes to complex environments.
It shows promise in ICRL as a meta-RL approach to learning effective corruption-robust algorithms.

\end{abstract}

% !TEX root =  main.tex
%%%%%%%%%%%%%%%%%%%%%%%%%%%%%%%%%%%%%
%%%%%%%%%%%%%%%%%%%%%%%%%%%%%%%%%%%%%
\section{Introduction}

% Important point: not single-task corruption, but an algorithm to produce a corrupt env

% \paulius{General background intro.}
Recent years have shown the impressive capabilities of transformer-based models on a range of tasks \citep{vaswani2017attention, raffel2020t5}.
The community has been shifting from single-task learning, to multi-task learning, and even multi-domain learning \citep{reed2022generalist}.
This has been made possible in part due to in-context learning, also called few-shot learning \citep{brown2020fewshot}, which allows models to adapt to new tasks simply by reading a handful of examples in the prompt, rather than requiring parameter updates. % or retraining.
Recently, transformers and in-context learning have found growing use in decision-making tasks, particularly in reinforcement learning (RL), where interactions with the environment replace traditional text-based examples \citep{chen2021decisiontransformer, xu2022promptingdt, laskin2022ad, lee2023dpt}.
In this paper, we focus on the robustness of in-context RL (ICRL) to reward poisoning attacks -- one of the major security threats for safe deployment of RL. 

% \goran{Intro to poisoning attacks.}
Reward poisoning attacks have been extensively explored in recent RL literature \citep{lin2017tactics, ma2019poisoning, zhang2020adaptive, wu2023reward, nika2023online}.
This line of work predominantly focuses on the canonical RL setting, modeling reward poisoning attacks as an attacker that corrupts the reward of a learning agent during training.
In contrast to \emph{test-time} adversarial attacks, poisoning attacks influence the policy that the agent adopts at test time; i.e., they are \emph{training-time} attacks.
This is perhaps not surprising, given that this line of work typically focuses on Markov stationary policies, implying that the agent's behavior is independent of the rewards at \emph{test time}.

% \goran{Novelty of our setting.}
However, an ICRL agent can implement a \emph{learning algorithm} in-context, using approaches such as Algorithm Distillation \citep{laskin2022ad} or the Decision-Pretrained Transformer \citep[DPT,][]{lee2023dpt}.
In this case, the context encodes past interactions (incl.\ rewards) between the environment and the agent. %, including past rewards.
By corrupting the agent's rewards at test time, an adversary can still influence the agent's behavior.
Simply put, such test-time reward poisoning schemes attack the learning algorithm implemented in-context. 

% \goran{Research question; A brief intro to DPT and perhaps some corruption robustness in RL.}
In this work, we aim to develop a novel training protocol for ICRL that enables models to be robust against test-time reward poisoning attacks.% targeting in-context learners.
We use DPT as a base approach.
At a high level, it should implement a corruption-robust learning algorithm in-context.
This differs from typical corruption-robust RL approaches \citep{lin2017tactics, ma2019poisoning, zhang2020adaptive, sun2021vulnerability, wu2023reward, nika2023online}: in our setting, rewards are corrupted only at test time, meaning the corruption does not affect the training process of the agent’s policy.
Our contributions are: % as follows.
% \goran{We are a bit shy when it comes to explaining our contributions}

\textbf{Framework.}
To our knowledge, we are the first to study reward poisoning in meta-RL -- we introduce and formalize a novel attack modality predicated on reward poisoning.
In this modality, the attacker influences the agent's context by corrupting the rewards. % that the agent's policy is conditioned on
More specifically, the attacker aims to minimize the agent's return by modifying the reward under a soft budget constraint encoded via a penalty term.

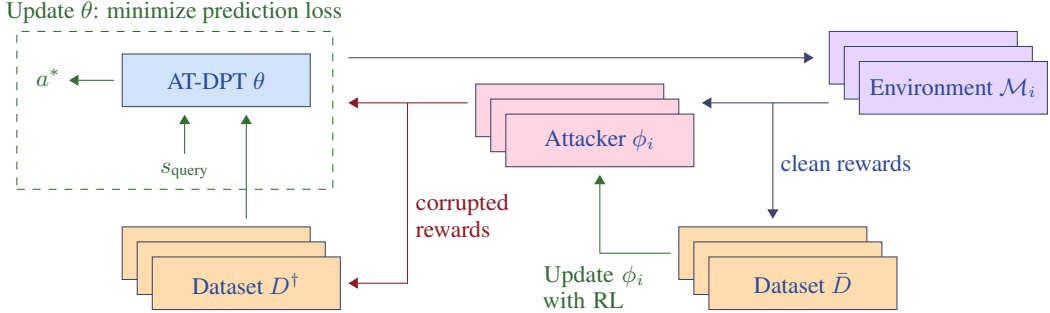
\begin{figure*}[t]
    \definecolor{mylightblue}{RGB}{210,227,255}
    \definecolor{myorange}{RGB}{254,220,174}
    \definecolor{myblue}{RGB}{43,74,153}
    % \definecolor{mygreen}{RGB}{61,103,62}
    \definecolor{mygreen}{RGB}{54, 107, 55}
    \definecolor{mypurple}{RGB}{230,214,255}
    \definecolor{myred}{RGB}{255,212,227}
    \definecolor{mydarkred}{RGB}{138,28,39}
    \definecolor{myborder}{RGB}{61,68,103}
    \centering
    {\footnotesize \begin{tikzpicture}[
        box/.style={draw=myborder, text=myblue, rectangle, minimum width=2.5cm, minimum height=7mm},
        arrow/.style={-{Triangle}, draw=myborder},
        every node/.style={anchor=west},
        node distance=0.25cm
    ]
        \node[draw=mygreen, dashed, rectangle, minimum width=4.2cm, minimum height=1.7cm, align=left, inner sep=2pt] at (-1.4,-0.4) {};
        \node[align=left,text=mygreen] at (-1.65, 0.9) {Update $\theta$: minimize prediction loss};

        \node[box,fill=mylightblue] at (0,0) (atdpt) {AT-DPT $\theta$};

        \draw[arrow,draw=mygreen] (atdpt.west) ++ (-0.1, 0) -- ++(-0.6, 0) node[left,yshift=0.4mm,text=mygreen] {$a^*$};
        \node[text=mygreen] (squery) at (0.4, -1.0) {$s_\text{query}$};
        
        \coordinate (atdptsouthgap) at ($(atdpt.south) + (0,-0.1)$);
        \draw[arrow,draw=mygreen] (squery.north) -- (squery |- atdptsouthgap);
        
        \node[box,fill=myorange] (d1) at (0.0,-1.9) {};
        \node[box,fill=myorange] (d2) at (0.2,-2.1) {};
        \node[box,fill=myorange] (d3) at (0.4,-2.3) {Dataset $D^\dagger$};

        \coordinate (d1northeastgap) at ($(d1.north) + (0.4, 0.1)$);
        \draw[arrow,draw=mygreen] (d1northeastgap) -- (d1northeastgap |- atdptsouthgap);

        \node[box,fill=mypurple] (env1) at (9.4, 0.3) {};
        \node[box,fill=mypurple] (env2) at (9.6, 0.1) {};
        \node[box,fill=mypurple] (env3) at (9.8, -0.1) {Environment $\Mcal_i$};
        
        \draw[arrow] (atdpt.east) ++ (0.5, 0.3) -- ($(env1.west) + (-0.2, 0)$);

        \node[box,fill=myred] (att1) at (4.7, -0.4) {};
        \node[box,fill=myred] (att2) at (4.9, -0.6) {};
        \node[box,fill=myred] (att3) at (5.1, -0.8) {Attacker $\phi_i$};

        \coordinate (env3westgap) at ($(env3.west) + (-0.4, -0.2)$);
        \coordinate (att3eastgap) at ($(att3.east) + (0.1, 0)$);
        \draw[arrow] (env3westgap) -- (env3westgap -| att3eastgap);

        \node[box,fill=myorange] (da1) at (7.4, -1.9) {};
        \node[box,fill=myorange] (da2) at (7.6, -2.1) {};
        \node[box,fill=myorange] (da3) at (7.8, -2.3) {Dataset $\Dbar$};
        
        \draw[arrow] (env3westgap -| da1) -- ($(da1.north) + (0, 0.1)$);
        \node[anchor=west,text=myblue] at (8.65, -0.9) {clean rewards};

        \coordinate (att1westgap) at ($(att1.west) + (-0.1, 0.1)$);
        \draw[arrow,draw=mydarkred] (att1westgap) -- ($(atdpt.east) + (0.5, -0.3)$);
        \draw[arrow,draw=mydarkred] ($(att1westgap) + (-0.8, 0)$) |- ($(d3.east) + (0.1, 0)$);
        
        \node[anchor=west,text=mydarkred,text width=1.8cm] at (3.8, -1.7) {corrupted rewards};

        \draw[arrow,draw=mygreen] (da1.west) ++ (-0.1, 0) -| ($(att3.south) + (0, -0.1)$);
        \node[text=mygreen,text width=1.8cm] at (5.5, -2.35) {Update $\phi_i$ with RL};
    \end{tikzpicture}}
    % \vspace{0.5em}
    \caption{
        The training procedure of AT-DPT.
        We use adversarial training to optimize the parameters $\theta$ of a transformer model, which is our learning agent.
        In each round, we first collect data by deploying the agent in $m$ environments.
        In environment $i \in {0, \dots, m}$, the agent observes rewards corrupted by an adversary defined by parameters $\phi_i$.
        We collect clean ($\Dbar$) and corrupted ($D^\dagger$) datasets containing trajectories with clean and corrupted rewards, respectively.
        The agent is trained to predict an optimal action for a query state given a context sampled from a corrupted dataset.
        The adversary is trained to minimize the agent’s return under a soft budget constraint, expressed as a penalty term. 
        During the test phase, the agent is deployed in a new corrupted environment.
    }
    \label{fig:setup}
\end{figure*}

\textbf{Method.}
We combine in-context learning with adversarial training to develop an agent that is robust against reward poisoning.
Specifically, we introduce the Adversarially Trained Decision-Pretrained Transformer (AT-DPT), which is trained by simultaneous optimization: a population of adversaries try to minimize the environment’s reward, while DPT learns to infer optimal actions from the corrupted data.
An overview of the training can be seen in \Cref{fig:setup}.
% training procedure

\textbf{Experiments.}
We conduct a systematic evaluation of the proposed method, comparing its corruption-robustness capabilities to various methods, including robust baselines designed to handle reward contamination, under various levels of poisoning.
Our results show that AT-DPT is robust against a wide range of reward poisoning attacks and yields better performance under corruption than the baselines. %considered.

\textbf{Importance.}
Prior works on reward poisoning attacks primarily focus on single-environment training-time attacks, focusing on learning policies.
In contrast, our work focuses on meta-RL, i.e., learning a learning algorithm itself.
Hence, our goal is to provide a principled approach to designing learning algorithms inherently robust to corruptions.
This can enable us to design more efficient algorithms that account for the distribution of tasks and utilize this information as a prior to obtain better performance.
% Yet, poisoning attacks in RL have primarily been considered standard RL settings.
Our results show the potential of transformer-based policies in implementing algorithms that are robust against data contamination.

\section{Related Work}\label{sec:related_work}

\textbf{Adversarial ML.}
Adversarial ML studies the effects of adversaries on models, as well as methods to defend against them.
These adversaries have been extensively studied in computer vision, and recently many have been proposed in RL, including test-time attacks on observations \citep{biggio2013evasion, szegedy2013intriguing, goodfellow2015adversarial, papernot2017practical}, (training-time) poisoning attacks \citep{mei2015using, li2016data, pattanaik2018robust}, and backdoor attacks \cite{chen2017targeted, gu2017badnets, salem2022dynamic}.
Closest to our work are poisoning attacks on RL \citep{huang2017adversarial, sun2021vulnerability, zhang2021robust}, which have considered different targets, including transitions \citep{ma2019poisoning, rakhsha2020policy}, rewards \citep{lin2017tactics, zhang2020adaptive, sun2020stealthy, wu2023reward, nika2023online}, states \citep{zhang2020robust}, and actions \citep{rangi2022understanding}.
Prior work also considered attacks in multi-agent RL \citep{mohammadi2023implicit, wu2024data}.
We contribute to this area by studying the robustness of ICRL under \emph{test-time} poisoning attacks.  

\textbf{Corruption Robustness in RL.}
% Prior work on robust RL defenses investigated methods against different types of adversarial perturbations -- against state perturbations 
% \citep{behzadan2017vulnerability, fischer2019online, zhang2020robust, kumar2022policy}, against reward \citep{ma2019poisoning, rakhsha2020policy, nika2023online} and transition perturbations \citep{zhang2021robust}.
% \goran{perhaps from this part we can focus on corruption robustness in RL}
Our work is closest to the works on corruption-robust bandits, RL, and multi-agent RL \citep{lykouris2018stochastic, rakhsha2020policy, niss2020crucb, chen2021improved, lee2021achieving, wei2022amodel, ding2022crlinucb, nika2023online, xu2024robustts}.
These works often establish guarantees for the suboptimality gap in terms of the level of corruption.
Rather than focusing on theory, we contribute a practical method for training corruption robust ICRL.
As explained in the introduction, this approach is conceptually different: corruption robust learning is implemented in-context.
We experimentally compare the efficacy of our approach to bandit and RL algorithms robust to reward contamination, such as corruption robust UCB \citep{niss2020crucb, ding2022crlinucb} and Natural Policy Gradient \citep[NPG,][]{kakade2001npg, zhang2021robust}.

Our work is also tied to the literature on robust offline RL \citep{yang2022rorl, panaganti2022robust, ye2023corruption, yang2023towards, zhihe2024dmbp}.
Prior work DeFog \citep{hu2023defog} or LHF \citep{chen2025filtering} rely on filtering the learning histories during training.
We note that both of these works are developed for robustness against random or noisy perturbations.
For \emph{adversarial} corruption robustness, another work \citep{xu2025robust} studies several improvements for the Decision Transformer.
However, this method focuses on the single-task setting, compared to ours.

% \goran{We don't need a motivating example in the related work}
% We motivate studying reward poisoning attacks by them being a simple enough framework with practical examples: in recommender (e.g., rating) systems -- an attack can possibly generate fake user profiles to tamper with ratings \citep{chen2022knowledge}; in language model preference datasets -- misspecified rewards of contaminated preference datasets, or an adversary interacting with a chatbot controlling the feedback provided to it \citep{zhang2021robust}.

% \goran{this paragraph is not needed?}
% Often these methods rely on adversarial training to acquire this robustness.
% We follow this line of work and consider a similar approach resembling GANs \citep{goodfellow2014gan} to study adversarial attacks against reward poisoning.

% \goran{remove this - it's not clear what we want to say..}
% We do not investigate any particular security concern, and present a method within a general framework; for a discussion of adversarial attacks in the context of practical security concerns we refer the reader to \citet{gilmer2018motivating}.
% gilmer2018motivating also has the physical knock-down-stop-sign attack

\textbf{In-Context Reinforcement Learning.}
Our research falls under ICRL, improving upon the initial DPT method \citep{lee2023dpt}.
A similar work is Algorithm Distillation, which distills a policy which implicitly imitates policy improvement by training on episodic trajectories from learning algorithm histories \citep{laskin2022ad}.
Follow-up work by \citet{zisman2024emergence} involves injecting noise in the curriculum to allow generating learning histories without the need for optimal actions.
However, both they and \citet{dong2024context} show that ICRL is sensitive to pretraining dataset perturbations.
We also mention \citet{tang2024adversarially}, who study the Adversarially Robust Decision Transformer (ARDT) -- a method robust against adaptive adversaries within a Markov game framework, capable of choosing actions which minimize the victim's rewards.
This framework, translated to ours, would correspond to an adversary modifying transition probabilities and the victim observing the adversary's action.
In contrast, we consider adversarial rewards generated by the attacker, and the victim only observes the realized reward without knowledge of whether an attacker is interfering, nor knowledge of their algorithm. % instead of adversarial transition probabilities,
For an in-depth discussion on ICRL, we refer to \citet{moeini2025survey}.

\textbf{Meta-RL.}
Our work is broadly related to meta-RL, since we consider a multi-task setting.
% Within decision-making and RL, 
Meta-RL has been used in a variety of ways -- optimizing a policy conditioned on histories of past transitions via an RNN \citep{duan2016rl2}, similarly, utilizing a Structured State Space Sequence model replacing the RNN \citep{lu2023s5}, learning good \textit{starting point} parameters that make learning in tasks faster \citep{finn2017model}, learning a dynamics model shared across tasks \citep{nagabandi2018learning}.
Transformers have also been utilized in prior work in learning multi-task policies \citep{reed2022generalist, lee2022multi}.
We refer the interested reader to \citet{beck2023survey} for a detailed discussion on meta-RL. 

\textbf{Other.} 
Recently there have been many works studying various different attacks on large language models (LLMs) to provoke an unsafe response \citep[and many others]{zhao2024iclbackdoor, he2024icl, cheng2024iclnoisy}, also called red-teaming \citep{ganguli2022redteaming}.
The increasing use of LLMs within decision-making systems provoke the need to study robustness.
Within the text domain \citet{cheng2024iclnoisy} find that pretraining a transformer with noisy labels robustifies it against that type of perturbation.
% We see that by training the DPT with poisoned rewards in the context leads to behavior that is robust against these perturbations.
Therefore, we advocate for the study of robust decision-making algorithms and hope our method contributes to this body of knowledge.

\section{Setup}\label{sec:setup}
\textbf{Notation.}
We will use $\Delta(\Acal)$ to refer to the probability distribution over $\Acal$, and $\norm{\,\cdot\,}_2$ denotes the Euclidean norm.
% We will use notation similar to \citet{lee2023dpt}.

\subsection{In-Context Sequential Decision-Making}

\textbf{Environment.} We consider a multi-task sequential decision making setting, where we denote $\Tcal$ as the distribution of tasks.
Each task $\Mcal \sim \Tcal$ is formalized as an episodic finite-horizon Markov decision process (MDP) ${\Mcal = \left< \Scal, \Acal, R, T, H, \rho \right>}$, where $\Scal$ is the state space, $\Acal$ is the action space, $R: \Scal \times \Acal \rightarrow \Delta(\R)$ is the reward function, $T: \Scal \times \Acal \rightarrow \Delta(\Scal)$ is the transition function, $H \in \N$ is the horizon, and $\rho \in \Delta(\Scal)$ is the starting state distribution.
We denote realized states, actions, and rewards at timestep $h$ by $s_h$, $a_h$, and $r_h$, respectively.
We distinguish between the clean and corrupted settings and use $\rbar_h$ to denote the true rewards, and $\rdagger_h$ to denote the rewards produced by an attacker.
The attack model is introduced in the next subsection.
Let $\mu_{\bar{R}}(s, a)$ denote the mean of the underlying environment reward for the state-action pair $(s,a)$.
For a stochastic policy $\pi: \Scal \rightarrow \Delta(\Acal)$, the value function is defined as ${V^\pi(\rho) = \E_{s \sim \rho} \left[\sum_{h=1}^{H} \rbar_h \mid \pi, s_0 = s\right]}$, where the expectation is w.r.t.\ the randomness of the underlying rewards when rolling out policy $\pi$ in $\Mcal$.
The solution to task $\Mcal$ is an optimal policy $\pi^\star_\Mcal$ that maximizes the value function, i.e., $V^{\pi_{\Mcal}^\star}(\rho) = \max_{\pi} V^{\pi}(\rho)$.

\textbf{Agent.}
We model a learning agent as a context-dependent policy parameterized by a transformer with parameters $\theta$ which maps the history of interactions $D$ and a query state $s_\text{query}$ to a distribution over actions.
We denote this policy by $\pi_{\theta}(a_h \mid D, s_h)$
%, where $\theta$ represents the parameters of a transformer model that implements the learning algorithm of the agent in-context
and $D = \{(s_i, a_i, r_i, s_{i+1})\}_{i=0}^{H-1}$ is the \emph{in-context dataset} consisting of a set of previous interactions.
% We follow the terminology by \citet{lee2023dpt} and refer to $D$ as the in-context dataset.
To implement an efficient learner, we can train $\pi_{\theta}(\,\cdot \mid D, s_h)$ to predict optimal actions $a_h^\star \sim \pi^\star_\Mcal(\,\cdot \mid s_h)$ for a task $\Mcal$ sampled from a given task distribution -- this approach is the backbone of the DPT \citep{lee2023dpt}.

\subsection{Attack Model}\label{sec:attack_model}
We consider bounded reward poisoning attacks applied to a fraction of tasks at \emph{test-time}.
We employ Huber's $\varepsilon$-contamination model \citep{huber1964contamination} and assume that the agent observes the corrupted reward in $\varepsilon$-fraction of timesteps.
We model the attacker $\pidagger_\phi: \Scal \times \Acal \times \R \times (\Scal \times \Acal \times \R \times \Scal)^C \rightarrow \Delta(\R)$ as a function of the state, action and reward of the last timestep along with an in-context dataset $\Bar{D} \in (\Scal \times \Acal \times \R \times \Scal)^C$ consisting of $C$ tuples of agent's interactions.
Formally, at timestep $h$, the environment generates $\rbar_h \sim R(s_h, a_h)$, and the agent observes
\begin{equation*}
    \tilde r_h = \begin{cases}
        \rdagger_h \sim \pidagger_\phi(\,\cdot \mid s_h, a_h, \rbar_h, \bar{D}) & \text{with probability }\varepsilon, \\
        \rbar_h & \text{otherwise.}%
    \end{cases}%
\end{equation*}
% \yigit{
% \begin{equation*}
%     \tilde r_h = \begin{cases}
%         \pidagger_\phi(\,\cdot \mid s_h, a_h, r_h), & \text{with probability }\varepsilon \\
%         \Bar{r}_h, & \text{otherwise}%
%     \end{cases}.
% \end{equation*}
% }
The attacker observes the underlying environment reward $\rbar_h$ to generate $\rdagger_h$, but the victim $\pi_\theta$ only observes the realized reward $\tilde{r}_h$.
We call an attacker \textit{adaptive} if $C > 0$, meaning it leverages the agent’s past interactions, and \textit{non-adaptive} if $C = 0$.
In the non-adaptive case ($C=0$) we simplify $\pidagger_\phi(\,\cdot \mid s_h, a_h, \rbar_h, \bar{D}) = \pidagger_\phi(\,\cdot \mid s_h, a_h, \rbar_h)$.
In both cases the attacker aims to minimize the agent's expected return in $\Mcal$ under a soft budget constraint, without forcing a specific policy for the agent.
We denote the mean and variance of corrupted rewards by $\mu_{\phi}(s,a) = \E_{r^\dagger \sim \pidagger_\phi(\,\cdot \mid s, a)}[r^\dagger]$ and $\sigma_\phi(s,a) = \V_{r^\dagger \sim \pidagger_\phi(\,\cdot \mid s, a)}[r^\dagger]$.
The attacker's objective is
\begin{equation}
\begin{split}
    L(\Mcal, \phi, \theta) &= \E \left[ \sum_{h=1}^H -\rbar_h  \mid \pi_\theta, \pidagger_\phi \right] \label{eq:attackers_objective} \\
     - \lambda \cdot c_\mu \big( &\norm{\mu_{\phi} - \mu_{\bar R}}_2 \big) - \lambda \cdot c_\sigma\left( \norm{\sigma_\phi}_2\right),
\end{split}
\end{equation}
% \begin{align*}
%     L(\Mcal, \phi, \theta) = \E \left[ \sum_{h=1}^H -\rbar_h  \mid \pi_\theta, \pidagger_\phi \right] - \lambda \cdot c\left (\norm{\mu_{\phi} - \mu_{\bar R}}_2^2 \right ),
% \end{align*}
where we take the expectation over the stochasticity of the environment, the agent's policy, and the contamination model.
% The expectation $L(\Mcal, \phi, \theta)$ is taken over the randomness of the realized rewards when running the policy $\pi_\theta$ in $\Mcal$, while corrupting its context $D$ using the $\varepsilon$-contamination model with the attack policy $\pidagger_{\phi}$.
% \yigit{
% $\lambda > 0$ is the penalty regularization coefficient and $c: \R \to \R$ is a penalty function for exceeding the attack budget $B$, e.g. $c(x) = \max(0, x - B)$.
% }
% The goal of the attacker is to minimize the agent's total return in $\Mcal$ under a soft-budget constraint on reward perturbations, expressed through a penalty term.
$c_\mu$, $c_\sigma$ are penalty functions for exceeding budget $B$ and $B_\sigma$ respectively, and $\lambda > 0$ controls the strength.
% In our experiments, we use $\norm{\cdot} \coloneqq \norm{\cdot}_{1, 1}$ where $\norm{A}_{1, 1} = \sum_{ij}|A_{ij}|$.
% We set $c_\mu(x) = c_\sigma(x) = \max(0, x - B)$ with $\lambda = 10$. % Moved to experiments 
% $\lambda > 0$, $c$ is a penalty function for exceeding budget $B$, and $\mu_{\phi} = \E_{r \sim \pidagger_\phi(\,\cdot \mid s, a)}[r]$ are the means of corrupted rewards.
% In our experiments, we use $c(x) = \max(0, x - B)$ and set $\lambda = 10$.
In both cases, we focus on non-behavior-targeted attacks (i.e., ones which do not force a specific policy), as opposed to behavior-targeted attacks, or policy-forcing attacks \citep{hussenot2019copycat, boloor2020attacking}.

\subsection{In-Context RL With Corrupted Rewards}\label{sec:equilibria}
% The attacker induces a change in \textit{perceived} rewards for the agent.
% To account for this, we modify the objective function of the agent to optimize its performance under corrupted feedback.
% More specifically, the objective of the agent is
% \goran{remove the previous two and put 'To account for this, we set the agent's objective to'}
To account for the change induced by the attacker, we set the agent's objective to 
$U(\Mcal, \theta, \phi) = \E \left[ \sum_h^H \rbar_h  \mid \pi_\theta, \pidagger_\phi \right]$,
where the expectation is taken over the randomness of the realized rewards when running the policy $\pi_\theta$ in $\Mcal$,
while corrupting its context $D^\dagger$ using the $\varepsilon$-contamination model with the attack policy $\pidagger_{\phi}$.

We search for a Nash equilibrium $(\theta^\star, \{\phi_{\Mcal}^\star\}_{\Mcal \in \Tcal})$ such that ${\theta^\star \in \argmax_{\theta} \E_{\Mcal \in \Tcal} [U(\Mcal, \theta, \phi^\star_{\Mcal})]}$ and ${\phi_{\Mcal}^\star \in \argmax_{\phi} L(\Mcal, \theta^\star, \phi)}$ for all $\Mcal \in \Tcal$. %\goran{Describe what we do in practice}.
% \paulius{you mean like describe supervised learning and REINFORCE? or do you mean describe that instead of argmax rewards, the agent considers argmin NLL(action)?}\goran{I mean there are many $\Mcal$, we can't ensure this holds for every $\Mcal$. I suppose you sample k $\Mcal$ and train k policies $\phi_{\Mcal}$. E.g., we could say 'We approximate this condition by sampling $k$ tasks $\Mcal$ and for each we optimize a separate policy $\pi_{\phi_{\mathcal_{M}}}$.'
% }
% To make this work in practice, we sample $M$ tasks $\{\Mcal_i \sim \Tcal\}_{i=0}^M$ and for every task $\Mcal_i$ we optimize a separate attacker .
% \begin{equation*}
%     (\theta^\star, \{\phi_{\Mcal}^\star\}_{\Mcal \in \Tcal}) \approx (\theta^\star, \{\phi_{\Mcal_i}^\star\}_i^M) ,
% \end{equation*}
% and by adversarially training DPT.
To approximate this equilibrium empirically, we propose an adversarial training procedure.
We sample $M$ tasks and optimize a dedicated attacker $\pidagger_{\phi_{\Mcal_i}}$ for each task $\Mcal_i$, concurrently with the optimization of the agent $\pi_\theta$.

\section{Method}\label{sec:method}

We extend the work by \citet{lee2023dpt} with adversarial training to mitigate attacks which poison the context and influence future actions.
The setup consists of three phases.

% Mądry et al. (2018) states that if an attack is robust to PGD adversaries, it would be robust against many different types of adversaries. In our case we show this principle extends to in-context learning -- the first row in Table 1 shows that AT-DPT can generalize to multiple different attacks while only being trained on one type.

\textbf{Pretraining.}
\citet{lee2023dpt} use GPT-2 as the underlying transformer model, and we adopt the same architecture.
The model $\pi_\theta$ is initialized from scratch and trained via supervised learning by predicting an optimal action from context $D_\text{pre}$ and query state $s_q$.
During the pretraining phase the model observes a dataset $D_{\text{pre}} \sim \mathcal{D}_{\text{pre}}$ which consists of tuples $(s, a, r, s')$ sampled from a set of $M$ tasks $\left\{ \Mcal_i \sim \Tcal \right\}_{i = 1}^M$.
This dataset can be collected in various ways, e.g., random interactions with the environments.
Alongside these interactions, we also sample a query state $s_q \sim \mathcal{D}_\text{query}$ and its corresponding optimal action $a^\star \sim \pi^\star_\Mcal(\,\cdot \mid s_q)$.
The model is then trained to minimize \(
    \min_\theta \E_{D_{\text{pre}} \sim \Dcal_{\text{pre}}, s_q \sim \mathcal{D}_\text{query}} \ell(\pi_\theta(\,\cdot \, | D_\text{pre}, s_q ), a^\star),
\) where $\ell$ is the NLL loss.

\textbf{In-Context Learning.}
During the test phase $\pi_\theta$ is deployed in $\Mcal \sim \Tcal$ with an empty context $D = \{\}$.
The original work updated the context $D$ with the entire trajectory $\left\{(s_h, a_h, r_h, s_{h+1})\right\}_{h=1}^H$ only after the entire episode \citep{lee2023dpt}.
Whereas, in our method we update context $D$ with transitions $(s_h, a_h, r_h, s_{h+1})$ after every timestep $h$, to support robustness against adaptive attacks.
\begin{algorithm}[t]
\caption{Adversarial Training for AT-DPT}
\label{alg:atdpt}
\begin{algorithmic}[1]
    \STATE \textbf{input:} victim $\pi_\theta$ -- DPT with pretrained params $\theta_0$
    \STATE \textbf{input:} attacker $\pidagger_\phi$ with initial params $\phi_0$, budget $B$, fraction of steps poisoned $\varepsilon$
    \STATE Sample $M$ tasks $\{ \Mcal_i \sim \Tcal \}_{i=1}^M$
    \FOR{round $n$ in $0 \dots N-1$, simultaneously in all $\Mcal$}
        \STATE roll out $\pi_{\theta_{n}}$ for $H$ steps in $\Mcal_i$ poisoned by $\pidagger_\phi$ with $\varepsilon$-contamination model and budget $B$,
            % \INDSTATE $a_h \sim \pi_{\theta_{n}} (\,\cdot \mid D, s_h)$
            % \INDSTATE $r_h = \begin{cases}
            %     \rdagger_h \sim \pi^\dagger_{\phi_n}(\,\cdot \mid s_h, a_h, \rbar_h) & \text{with prob.\ } \varepsilon \\
            %     \rbar_h \sim R(\,\cdot \mid s_h, a_h) & \text{otherwise}
            % \end{cases}$
            % \INDSTATE $s_{h+1} \sim T(\,\cdot \mid s_h, a_h)$
            %
            % \INDSTATE \goran{
            % % I'd also explicitly state (e.g.: apply $\epsilon$-contamination model with $\pi_{\phi_n}$ to obtain (corrupted rewards) $\tilde r_h$).
            % Is it easy to say that we mean by the points below?
            % e.g., collect corrupted dataset $D$ for DPT? and collect dataset $\Dbar$ for the attacker?
            % } 
            \INDSTATE where DPT collects corrupted dataset $D^\dagger$, and attacker collects dataset $\Dbar$% for the attacker
        \STATE $\phi_{n+1} \gets$ train on $\Dbar$ with RL:
        % \INDSTATE $\min_\phi \sum^H_h \bar{r}_h$, see \Cref{eq:attackers_objective} %s.t.~$c\left (\norm{\mu_{\phi} - \mu_{\bar R}}_2 \right ) \leq B$
        \INDSTATE see \Cref{eq:attackers_objective}
        \STATE $\theta_{n+1} \gets$ train on $D^\dagger$ via supervised learning:
        \INDSTATE $\min_\theta \ell(\pi_\theta(\,\cdot \mid D^\dagger, s_q), a^\star)$, $a^\star$ provided by oracle \\
    \ENDFOR
\end{algorithmic}
\end{algorithm}

\textbf{Adversarial Training.}
Before testing, we include an additional phase for adversarial training.
An illustration of this training process is shown in \Cref{fig:setup}.
In the adversarial setting $\pi_\theta$ is deployed in $\Mcal$ under an attacker $\pidagger_\phi$, contaminating the victim's dataset $D^\dagger$ as specified in the previous section.
% Recall the rewards $r_h$ observed by DPT in the context $D$ may be poisoned.
We account for this by introducing an additional adversarial training stage between the original pretraining and in-context learning.
% In this stage we collect additionally contexts for training $\pi_\theta$ and $\pidagger_\phi$ by rolling out policy $\pi_\theta$ in $M$  sampled poisoned tasks $A_\phi(\Mcal_i), i\in[M]$.
To train the agent and the attacker, recall that we use two different contexts -- a context with poisoned rewards $D^\dagger$ for the agent, and a context with underlying rewards $\Dbar$ for the attacker.
% \goran{where is this context?}
We repeat this process for $N$ rounds, updating $\theta$ and $\phi$ after each round.
Parameters $\theta$ are updated as in the original DPT setting, with $s_q$ sampled from the environment, and $a^\star$ provided by an oracle.\footnote{In the algorithm we require access to clean environments sampled from $\Tcal$ at training time, although offline trajectories could be used with simulated attacks and (near-)optimal actions.}
The pseudocode of this method can be found in \Cref{alg:atdpt}.

We consider attackers parameterized by $\phi$ (e.g., a neural network).
To train the attacker we use the REINFORCE algorithm \citep{williams1992reinforce} -- after each episode we update $\phi$ with the objective specified in \Cref{eq:attackers_objective}.
Recall that while the victim $\pi_\theta$ only observes the realized reward $\tilde{r}_h$, the attacker has to have access to the underlying environment reward $\rbar_h$.
The attacker's goal is to poison a single algorithm, which we denote the \textbf{attacker target}.
That is, a different policy might emerge from an attacker targeting DPT versus an attacker targeting TS.

\textbf{Bandit Setting.}
In the bandit settings we consider a direct parameterization of a deterministic attack, i.e., for an action $a_h^{(i)}$ (choosing arm $i$) at timestep $h$ the attack is $\pidagger_\phi(\,\cdot \mid a_h^{(i)}, \rbar_h) = \pidagger_\phi(a_h^{(i)}, \rbar_h) = \rbar_h + \phi(i)$, where $\phi \in \R^{|\Acal|}$.
% The budget $B$ constraint then becomes\todo{check under new formulation} $\left\| \phi \right\|_2 < B$, which is incorporated into the REINFORCE objective.

\textbf{Adaptive Attacker.}
We also consider a context-dependent algorithm, e.g., a transformer, to enable the attacker to adapt to the defenses of the victim.
% \todo{check if this is already defined in the new formulation}
For this we utilize the same architecture (GPT-2) as the victim.
The interaction in the environment is then modified as follows.
At the start of an episode empty context $\Dbar = \{\}$ is initialized for the attacker.
At every step $h$ the attacker samples a reward $\rdagger_h \sim \pidagger_\phi(\,\cdot \mid \Dbar, s_h, a_h, \rbar_h)$ for the victim and appends $(s_h, a_h, \rbar_h)$ to the dataset $\Dbar$.
% At the end of each round we train the attacker via REINFORCE to minimize underlying rewards $\rbar$ received by victim.

\textbf{MDP Setting.}
In the MDP setting we also consider a direct parameterization of a deterministic non-adaptive attack, i.e., for a state-action pair $(s^{(i)}, a^{(j)})$ the attack becomes ${\pidagger_\phi(\,\cdot \mid s^{(i)}, a^{(j)}, \rbar) = \pidagger_\phi(s^{(i)}, a^{(j)}, \rbar) = \rbar + \phi(i, j)}$, where $\phi \in \R^{|\Scal| \times |\Acal|}$. % and, similarly, the constraint $\left\| \phi \right\|_2 < \delta$ is incorporated into the objective.
% !TEX root =  main.tex
%%%%%%%%%%%%%%%%%%%%%%%%%%%%%%%%%%%%%
%%%%%%%%%%%%%%%%%%%%%%%%%%%%%%%%%%%%%
\section{Experiments}\label{sec:experiments}

We sample $M = 200$ tasks to run in parallel.
For each round we train both the attacker and DPT for multiple (e.g., 20) iterations on the same dataset.
% We use a transformer with 4 layers, 4 attention heads per layer, without dropout, and an embedding dimension of size 32.
% We set the context length equal to the horizon length $H=500$.
% For pretraining we use a learning rate of $0.001$, and during adversarial training we use $0.0001$.
We set penalties for exceeding the budget $c_\mu(x; B) = \max(0, x - B)$ and $c_\sigma(x; B_\sigma) = \max(0, x - B_\sigma)$ with $\lambda = 10$.
% The implementation of AT-DPT is provided in the supplementary material.
In all tables results are color coded using a gradient from orange (best) to white (worst), with color intensity decreasing as performance decreases (i.e., higher regret or lower reward).

\NewDocumentCommand{\tikzcellcolor}{O{19mm} O{0.85mm} m m}{%
    \setsepchar[.]{ }%
    % \readlist\seperatedval{#4}%
    \begin{tikzpicture}[remember picture, baseline,every node/.style={inner ysep=#2, inner xsep=-1mm,outer sep=0}]
    {\footnotesize
        \pgfmathsetlength{\textheight}{\dimexpr\ht\strutbox+\dp\strutbox\relax} % Ascender + Descender
        % \node[fill=#3, text width=#1, align=center, yshift=0.9mm] (A) {\phantom{\seperatedval[2]}};
        \node[anchor=south west, fill=#3, text width=#1, align=center] at (0, -1mm) (A) {#4};
        % \node[anchor=east, left=-9.5mm of A.west] {\seperatedval[1]};
        % \node[anchor=center] at (A) {\;\;\,\seperatedval[2]};
        % \node[anchor=west, right=-6mm of A.east] {\seperatedval[3]};
    }
    \end{tikzpicture}
}

\NewDocumentCommand{\colorizeValue}{O{19mm} O{0.85mm} m m m}{%
    \IfSubStr{#5}{N/A}{#5}{%
        \pgfmathsetmacro{\minvalue}{#3}%
        \pgfmathsetmacro{\maxvalue}{#4}%
        \def\textvalue{#5}%
        \setsepchar[.]{ }%
        \readlist\seperatedval{#5}%
        \def\value{\seperatedval[1]}%
        \pgfmathsetmacro{\value}{\seperatedval[1]}%
        \ifdim\value pt<\minvalue pt%
            \;\;\tikzcellcolor[#1][#2]{orange!55} \textvalue%
        \else\ifdim\value pt>\maxvalue pt%
            \;\;\tikzcellcolor[#1][#2]{orange!5} \textvalue%
        \else
            \pgfmathsetmacro{\colorvalue}{(1 - (\value - \minvalue) / (\maxvalue - \minvalue)) * 45 + 5}%
            \pgfmathsetmacro{\roundedcolorvalue}{int(\colorvalue)}%
            \;\;\tikzcellcolor[#1][#2]{orange!\roundedcolorvalue}{\textvalue}%
        \fi\fi%
    }%
}

\NewDocumentCommand{\colorizeValueMax}{O{18mm} O{0.85mm} m m m}{%
    \IfSubStr{#5}{N/A}{#5}{%
        \pgfmathsetmacro{\minvalue}{#3}%
        \pgfmathsetmacro{\maxvalue}{#4}%
        \def\textvalue{#5}%
        \setsepchar[.]{ }%
        \readlist\seperatedval{#5}%
        \def\value{\seperatedval[1]}%
        \pgfmathsetmacro{\value}{\seperatedval[1]}%
        \ifdim\value pt<\minvalue pt%
            \;\;\tikzcellcolor[#1][#2]{orange!5} \textvalue%
        \else\ifdim\value pt>\maxvalue pt%
            \;\;\tikzcellcolor[#1][#2]{orange!55} \textvalue%
        \else
            \pgfmathsetmacro{\colorvalue}{(\value - \minvalue) / (\maxvalue - \minvalue) * 45 + 5}%
            \pgfmathsetmacro{\roundedcolorvalue}{int(\colorvalue)}%
            \;\;\tikzcellcolor[#1][#2]{orange!\roundedcolorvalue}{\textvalue}%
        \fi\fi%
    }%
}

\newcommand{\colorBA}[6]{%
  \colorizeValue[18mm][0.8mm]{10}{30}{#1} &
  \colorizeValue[18mm][0.8mm]{10}{30}{#2} &
  \colorizeValue[18mm][0.8mm]{10}{30}{#3} &
  \colorizeValue[18mm][0.8mm]{10}{30}{#4} &
  \colorizeValue[18mm][0.8mm]{10}{30}{#5} &
  \colorizeValue[18mm][0.8mm]{10}{30}{#6}%
}
\newcommand{\colorBAclean}[2]{%
  \colorizeValue[18mm][0.8mm]{9}{25}{#1} &
  \colorizeValue[18mm][0.8mm]{9}{25}{#2}%
}

\newcommand{\colorBB}[6]{%
  \colorizeValue[18mm][0.8mm]{20}{60}{#1} &
  \colorizeValue[18mm][0.8mm]{20}{60}{#2} &
  \colorizeValue[18mm][0.8mm]{20}{60}{#3} &
  \colorizeValue[18mm][0.8mm]{20}{60}{#4} &
  \colorizeValue[18mm][0.8mm]{20}{60}{#5} &
  \colorizeValue[18mm][0.8mm]{20}{60}{#6}%
}
\newcommand{\colorBBclean}[2]{%
  \colorizeValue[18mm][0.8mm]{10}{30}{#1} &
  \colorizeValue[18mm][0.8mm]{10}{30}{#2}%
}

\newcommand{\colorBC}[6]{%
  \colorizeValue[17mm][0.8mm]{40}{100}{#1} &
  \colorizeValue[17mm][0.8mm]{40}{100}{#2} &
  \colorizeValue[17mm][0.8mm]{40}{100}{#3} &
  \colorizeValue[17mm][0.8mm]{40}{100}{#4} &
  \colorizeValue[17mm][0.8mm]{40}{100}{#5} &
  \colorizeValue[17mm][0.8mm]{40}{100}{#6}%
}
\newcommand{\colorBCclean}[2]{%
  \colorizeValue[17mm][0.8mm]{10}{70}{#1} &
  \colorizeValue[17mm][0.8mm]{10}{50}{#2}%
}

\newcommand{\colorBAdaptiveVsNonA}[6]{%
  \colorizeValue[19mm][0.8mm]{20}{90}{#1} &
  \colorizeValue[19mm][0.8mm]{20}{90}{#2} &
  \colorizeValue[19mm][0.8mm]{20}{50}{#3} &
  \colorizeValue[19mm][0.8mm]{20}{50}{#4} &
  \colorizeValue[19mm][0.8mm]{10}{40}{#5} &
  \colorizeValue[19mm][0.8mm]{10}{40}{#6}%
}
\newcommand{\colorBAdaptiveVsNonB}[6]{%
  \colorizeValue[19mm][0.8mm]{30}{100}{#1} &
  \colorizeValue[19mm][0.8mm]{30}{100}{#2} &
  \colorizeValue[19mm][0.8mm]{30}{70}{#3} &
  \colorizeValue[19mm][0.8mm]{30}{70}{#4} &
  \colorizeValue[19mm][0.8mm]{10}{40}{#5} &
  \colorizeValue[19mm][0.8mm]{10}{40}{#6}%
}
\newcommand{\colorBAdaptiveVsNonC}[6]{%
  \colorizeValue[19mm][0.8mm]{40}{100}{#1} &
  \colorizeValue[19mm][0.8mm]{40}{100}{#2} &
  \colorizeValue[19mm][0.8mm]{40}{100}{#3} &
  \colorizeValue[19mm][0.8mm]{40}{100}{#4} &
  \colorizeValue[19mm][0.8mm]{10}{70}{#5} &
  \colorizeValue[19mm][0.8mm]{10}{50}{#6}%
}

\newcommand{\colorBAdaptiveA}[6]{%
  \colorizeValue[20mm][0.8mm]{10}{50}{#1} &
  \colorizeValue[20mm][0.8mm]{10}{50}{#2} &
  \colorizeValue[20mm][0.8mm]{10}{50}{#3} &
  \colorizeValue[20mm][0.8mm]{10}{50}{#4} &
  \colorizeValue[20mm][0.8mm]{10}{50}{#5} &
  \colorizeValue[20mm][0.8mm]{10}{50}{#6}%
}
\newcommand{\colorBAdaptiveB}[6]{%
  \colorizeValue[20mm][0.8mm]{10}{70}{#1} &
  \colorizeValue[20mm][0.8mm]{10}{70}{#2} &
  \colorizeValue[20mm][0.8mm]{10}{70}{#3} &
  \colorizeValue[20mm][0.8mm]{10}{70}{#4} &
  \colorizeValue[20mm][0.8mm]{10}{50}{#5} &
  \colorizeValue[20mm][0.8mm]{10}{50}{#6}%
}
\newcommand{\colorBAdaptiveC}[6]{%
  \colorizeValue[20mm][0.8mm]{60}{80}{#1} &
  \colorizeValue[20mm][0.8mm]{60}{80}{#2} &
  \colorizeValue[20mm][0.8mm]{60}{80}{#3} &
  \colorizeValue[20mm][0.8mm]{60}{80}{#4} &
  \colorizeValue[20mm][0.8mm]{10}{50}{#5} &
  \colorizeValue[20mm][0.8mm]{10}{50}{#6}%
}

\newcommand{\colorMC}[6]{%
  \colorizeValueMax[21mm]{190}{260}{#1} &
  \colorizeValueMax[21mm]{190}{260}{#2} &
  \colorizeValueMax[21mm]{190}{260}{#3} &
  \colorizeValueMax[21mm]{190}{260}{#4} &
  \colorizeValueMax[21mm]{190}{260}{#5} &
  \colorizeValueMax[21mm]{190}{260}{#6}%
}

\newcommand{\colorLBC}[6]{%
  \colorizeValue[21mm]{31}{45}{#1} &
  \colorizeValue[21mm]{31}{45}{#2} &
  \colorizeValue[21mm]{31}{45}{#3} &
  \colorizeValue[21mm]{31}{45}{#4} &
  \colorizeValue[21mm]{2.9}{12}{#5} &
  \colorizeValue[21mm]{2.9}{9}{#6}%
}

\newcommand{\colorALBC}[8]{%
  \colorizeValue[21mm]{20}{100}{#1} &
  \colorizeValue[21mm]{20}{100}{#2} &
  \colorizeValue[21mm]{20}{100}{#3} &
  \colorizeValue[21mm]{20}{100}{#4} &
  \colorizeValue[21mm]{20}{100}{#5} &
  \colorizeValue[21mm]{20}{100}{#6} &
  \colorizeValue[21mm]{5}{50}{#7} &
  \colorizeValue[21mm]{5}{50}{#8}%
}

\newcommand{\colorMw}[4]{%
  \colorizeValueMax[21mm]{78}{120}{#1} &
  \colorizeValueMax[21mm]{78}{120}{#2} &
  \colorizeValueMax[21mm]{78}{120}{#3} &
  \colorizeValueMax[21mm]{78}{125}{#4}%
}

\newcommand{\colorMwA}[4]{%
  \colorizeValueMax[21mm]{65}{125}{#1} &
  \colorizeValueMax[21mm]{65}{125}{#2} &
  \colorizeValueMax[21mm]{65}{125}{#3} &
  \colorizeValueMax[21mm]{65}{125}{#4}%
}

\subsection{Baseline Algorithms}
To evaluate our method's performance in the bandit setting we compare it with widely used baselines:
Thompson sampling \citep[TS,][]{thompson1933sampling},
upper confidence bound \citep[UCB,][]{auer2002ucb}, and corruption robust algorithms: robust Thompson sampling \citep[RTS,][]{xu2024robustts} -- a TS-based algorithm robust to adversarial reward poisoning, which features an added term to the bonus term in TS, 
and corruption-robust upper confidence bound \citep[crUCB,][]{niss2020crucb} -- a UCB style algorithm robust to $\varepsilon$-contamination, where we chose the trimmed mean variant (the mean is estimated with a fraction of smallest and largest observed values removed for every arm).
For linear bandits we compare our method to LinUCB \citep{li2010linucb}, and a corruption robust variant -- CRLinUCB \citep[Section 4]{ding2022crlinucb}.

For the MDP baselines we choose two policy-gradient based methods -- natural policy gradient \citep[NPG,][]{kakade2001npg}; proximal policy optimization \citep[PPO,][]{schulman2017ppo}; and a value-based method -- Q-learning \citep{watkins1992qlearning}.
Additionally, we include DPT with frozen parameters (indicated as DPT) as a baseline to observe the effect of adversarial training, and we additionally include results with suboptimal demonstrations -- rows indicated by AT-DPT (sub.\ 30\%) mean a random 30\% of timesteps the expert actions are replaced with the 2nd best arm.
More details about the baselines can be found in \Cref{app:baselines}.

We note that the literature usually discusses theoretical algorithms which do not have practical implementations nor empirical results \citep{lykouris2018stochastic,lykouris2021corruption,chen2021improved}.
Given their principled nature it would be interesting to include them for comparison, but we believe it is outside the scope of this work to implement and tune them.

In addition to algorithm baselines, we also consider two baselines for evaluation.
We show performance of the algorithms in the clean environment, and we also consider a uniform random attack -- the poisoned reward for timestep $h$ is $\rdagger_h = \rbar_h + \phi(i)$, where $\phi \in \R^{|\Acal|}$ is generated once at the start of evaluation by sampling from a uniform random distribution and clipped by the budget constraint $\norm{\phi}_2 < B$.

\subsection{Bandit Setting}

% !TEX root =  main.tex
%%%%%%%%%%%%%%%%%%%%%%%%%%%%%%%%%%%%%
%%%%%%%%%%%%%%%%%%%%%%%%%%%%%%%%%%%%%

% \goran{Might be good to add paragraph heading for important paragraph. E.g., here we could put 'Environments.' See my comment below as well. The same for the next subsection.}
\textbf{Environment.}
We begin with empirical results in a simple scenario -- the multi-armed bandit problem.
%, which can be viewed as an MDP containing just one state ($\Scal = \{ s_0 \}$).
We follow a similar bandit setup to that presented in the original DPT paper \citep{lee2023dpt}.
We sample 5-armed bandits ($|\Acal| = 5$), each arm's reward function being a normal distribution $R(\,\cdot \mid s, a) = \Ncal(\mu^{(a)}, \sigma^2)$, where $\mu^{(a)} \sim \text{Unif}[0, 1]$ independently and $\sigma = 0.3$.
The optimal policy in this environment is to always choose the arm with the largest mean: $a^\star = \argmax_a \mu^{(a)}$.
We follow the same pretraining scheme as the original work.
For evaluation, we present the empirical cumulative regret: $\sum_h \rbar({a^\star}) - \rbar({a_h})$.
Low regret indicates the policy is close to optimal.
% Further details and hyperparameters used are shown in \Cref{app:experiments}.
%\goran{The statement doesn't seem to be specific to the paragraph below. Perhaps we can put 'Training curves'?}
%\paulius{I haven't made any training curves besides figure 1. Should I make more of them?}

\setlength{\tabcolsep}{-2.5pt}
\begin{table*}[!b]
    \caption{
        Comparison of cumulative regret (lower is better) of different algorithms under different attackers trained for 20 rounds, with $\varepsilon=0.4$ steps poisoned. For $\varepsilon=0.1$ and $0.2$ see \Cref{tab:bandit_megatable_epse0.8_appendix} in the Appendix.
        Mean and 95\% confidence interval (2$\times$SEM) over 10 runs.
        Attack budget $B = 3$. 
        $^*$ We use tuned versions of RTS and crUCB which outperform the base versions; full comparisons including base versions are given in \Cref{app:baselines}.
    }
    % \vspace{0.4em}
    \hspace*{-0.1em}
    \centering
    \begin{footnotesize}
    \centering
    \begin{tabular}{lcccccc|cc}
        \toprule
        \multirow{2}{*}{Algorithm} & \multicolumn{6}{c|}{Attacker Target} & \multirow[b]{2}{*}{\shortstack[c]{Unif.\ Rand.\\Attack}} & \multirow{2}{*}{Clean Env.} \\
        & AT-DPT & DPT & TS & RTS$^*$ & UCB1.0 & crUCB$^*$ \\
        \midrule
        AT-DPT                & \colorBC{24.2 $\pm$ 1.2}{24.8 $\pm$ 1.4}{29.8 $\pm$ 3.0}{28.3 $\pm$ 1.9}{24.5 $\pm$ 0.8}{23.8 $\pm$ 1.4} &  \colorBCclean{38.7 $\pm$ 1.7}{13.0 $\pm$ 0.9} \\
        AT-DPT (sub.\ 30\%)\ \  & \colorBC{41.2 $\pm$ 2.9}{41.9 $\pm$ 3.8}{45.7 $\pm$ 3.3}{43.6 $\pm$ 2.5}{41.6 $\pm$ 3.0}{41.0 $\pm$ 3.4} &  \colorBCclean{47.8 $\pm$ 3.5}{25.9 $\pm$ 3.3} \\
        DPT                   & \colorBC{63.6 $\pm$ 8.6}{59.4 $\pm$ 5.2}{62.0 $\pm$ 8.6}{59.1 $\pm$ 7.3}{55.4 $\pm$ 8.1}{58.8 $\pm$ 7.8} &  \colorBCclean{37.2 $\pm$ 1.2}{11.5 $\pm$ 0.5} \\
        TS                    & \colorBC{106.3 $\pm$ 3.8}{97.7 $\pm$ 4.9}{94.3 $\pm$ 3.8}{93.1 $\pm$ 6.0}{89.6 $\pm$ 1.8}{92.6 $\pm$ 4.8} & \colorBCclean{34.2 $\pm$ 1.6}{8.7 $\pm$ 0.6} \\
        RTS$^*$               & \colorBC{102.9 $\pm$ 4.5}{97.0 $\pm$ 4.2}{90.4 $\pm$ 4.4}{92.5 $\pm$ 5.0}{89.2 $\pm$ 3.4}{89.0 $\pm$ 3.0} & \colorBCclean{33.9 $\pm$ 1.6}{10.2 $\pm$ 0.4} \\
        UCB1.0                & \colorBC{104.1 $\pm$ 3.6}{95.8 $\pm$ 4.9}{90.6 $\pm$ 4.1}{90.0 $\pm$ 5.2}{88.1 $\pm$ 3.4}{91.2 $\pm$ 3.4} & \colorBCclean{38.1 $\pm$ 2.2}{16.0 $\pm$ 0.5} \\
        crUCB$^*$             & \colorBC{86.0 $\pm$ 4.4}{85.0 $\pm$ 2.3}{82.0 $\pm$ 4.4}{82.4 $\pm$ 3.3}{79.4 $\pm$ 3.0}{82.5 $\pm$ 5.1} &  \colorBCclean{31.8 $\pm$ 1.6}{15.8 $\pm$ 0.5} \\
        \bottomrule
    \end{tabular}
    \end{footnotesize}
    \label{tab:bandit_megatable_epse0.8}
    \vspace*{-0.5em}
\end{table*}

\textbf{Hyperparameters}
To pretrain DPT in the bandit setting we use the following architecture and hyperparameters.
The Transformer has 4 layers, 4 attention heads per layer, embedding dimension -- 32, no dropout.
We set the context length equal to episode length $H=500$, learning rate $\eta = 0.001$ and train for $400$ epochs.
We pretrain DPT in the same way as the original, see the work by \citet{lee2023dpt} for more details.
For adversarial training we use $\eta=0.0001$ for the victim and $\eta_\text{attacker}=0.03$.
We consider attackers with a diagonal covariance matrix, and set $B_\sigma=1$.

\begin{figure}[t]
    \begin{center}
    \centerline{\includegraphics[width=\columnwidth]{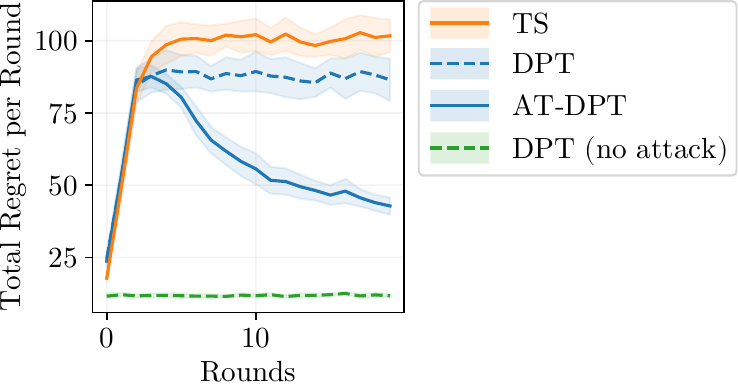}}
    % \vspace{0.7em}
    \caption{
        Comparison of the cumulative regret per round (lower is better) of different methods throughout 20 rounds of adversarial training (simultaneously learning AT-DPT and attackers) in the bandit setting.
        Within one round we perform $H = 500$ steps.
        The y axis indicates total regret for that round.
        Mean and 95\% confidence interval (2$\times$SEM) over 10 experiment replications.
        Attack budget $B = 3$, $\varepsilon = 0.4$.
    }
    \label{fig:adv_bandit_dpt}
    \end{center}
    \vspace*{-2em}
\end{figure}

\textbf{Adversarial training makes DPT robust to poisoning attacks.}
In \Cref{fig:adv_bandit_dpt}, we present the training-time performance of DPT under adversarial training.
The training curve shows the per-round cumulative regret, averaged across $M$ tasks seen during training.
We observe the regret significantly increase in the first rounds, but in further rounds DPT learns to recover from the attacks, and results improve.
In the figure we compare performance of TS and frozen DPT under the same attack, and also show the performance of frozen DPT on the clean environment (no attack).
We refer to the adversarially trained DPT models as AT-DPT. 

\setlength{\tabcolsep}{0pt}
{
\setlength{\textfloatsep}{5pt}
\begin{table*}[t]
    \caption{Comparison of the cumulative regret (lower is better) of adaptive and non-adaptive attackers. Attackers trained for 400 rounds, with $\varepsilon=0.4$ steps poisoned. For $\varepsilon=0.1$ and $0.2$ see \Cref{tab:bandit_adaptive_full} in the Appendix.
    AT-DPT (A) means AT-DPT trained against the adaptive attacker, AT-DPT (n-A) means AT-DPT trained against the non-adaptive attacker.
    Mean and 95\% confidence interval (2$\times$SEM) over 10 replications. Attack budget $B = 3$. 
    $^*$ We use tuned versions of RTS and crUCB which outperform base versions; details in \Cref{app:baselines}.
    }
    % \vspace{0.4em}
    \footnotesize
    \centering
    \begin{tabular}{lcc|cc|cc}
        \toprule
        \multirow{3}{*}{Algorithm} & \multicolumn{4}{c|}{Attacker Target} & \multirow[b]{3}{*}{\shortstack[c]{Unif.\ Rand.\\Attack}} & \multirow{3}{*}{Clean Env.} \\
        & \multicolumn{2}{c|}{Adaptive} & \multicolumn{2}{c|}{Non-adaptive} \\
        & AT-DPT & TS & AT-DPT & TS & \\
        \midrule
        AT-DPT (A)                    & \colorBAdaptiveVsNonC{37.1 $\pm$ 6.6}{36.4 $\pm$ 9.4}{38.0 $\pm$ 6.4}{42.6 $\pm$ 6.7    }{41.4 $\pm$ 7.3}{21.3 $\pm$ 9.0} \\
        AT-DPT (n-A)                  & \colorBAdaptiveVsNonC{88.1 $\pm$ 20.0}{81.0 $\pm$ 11.2}{22.8 $\pm$ 1.6}{29.8 $\pm$ 2.2  }{39.7 $\pm$ 3.8}{13.8 $\pm$ 1.2} \\
        DPT                           & \colorBAdaptiveVsNonC{97.9 $\pm$ 18.6}{82.1 $\pm$ 20.7}{61.6 $\pm$ 8.0}{61.6 $\pm$ 6.6  }{37.3 $\pm$ 3.5}{12.1 $\pm$ 0.8} \\
        TS                            & \colorBAdaptiveVsNonC{90.2 $\pm$ 21.9}{104.2 $\pm$ 26.7}{106.3 $\pm$ 5.5}{94.3 $\pm$ 4.8}{34.1 $\pm$ 2.5}{9.1 $\pm$ 0.7} \\
        RTS$^*$                       & \colorBAdaptiveVsNonC{90.5 $\pm$ 21.3}{103.6 $\pm$ 26.8}{104.5 $\pm$ 5.5}{90.9 $\pm$ 4.2}{34.5 $\pm$ 2.4}{10.5 $\pm$ 0.6} \\
        UCB                           & \colorBAdaptiveVsNonC{94.3 $\pm$ 22.4}{103.9 $\pm$ 28.4}{101.3 $\pm$ 5.0}{87.8 $\pm$ 4.4}{38.2 $\pm$ 1.6}{16.0 $\pm$ 0.4} \\
        crUCB$^*$                     & \colorBAdaptiveVsNonC{85.1 $\pm$ 23.5}{79.6 $\pm$ 29.4}{88.4 $\pm$ 4.4}{79.9 $\pm$ 4.7  }{32.0 $\pm$ 1.7}{15.8 $\pm$ 0.3} \\
        \bottomrule
    \end{tabular}
    \label{tab:bandit_adaptive_motivation}
\end{table*}
}

\textbf{Evaluation.}
To evaluate AT-DPT on attacks trained for it, we cross-validate to prevent evaluation on the same attack AT-DPT has seen during training -- we evaluate one AT-DPT with an attacker which is targeting AT-DPT for a different seed.
We report the mean and 95\% confidence interval (2$\times$SEM) across 10 different experiment replications.
During the test phase, AT-DPT uses trained parameters $\theta$, and the attacker uses $\phi$.
The procedure for this can be seen in Appendix \Cref{alg:atdpt-deployment}.
We note, that the tables and plots show the performance based on clean rewards, and not $\rtilde$.
We run adversarial training for $N=20$ rounds. %, with varying fractions $\varepsilon$ of steps poisoned.
\Cref{tab:bandit_megatable_epse0.8} presents an extensive evaluation of AT-DPT and other method performance against attackers targeting different methods.
We can clearly see AT-DPT outperforming all baselines in an adversarial setting.
Given that AT-DPT displays robustness against different attackers, it illustrates that it can successfully recover from attacks that are out-of-distribution.
% Although, adversarial training seems to trade-off the performance in the clean and random attack environment, where the frozen model (DPT), or even a baseline algorithm like TS perform better.

\textbf{Adaptive Attacks.}
For the adaptive attacker we utilize the same architecture as the victim, except without pretraining.
In this setting we use $\eta_\text{attacker}=0.00003$.
\Cref{tab:bandit_adaptive_motivation} shows a comparison of performance under both adaptive and non-adaptive attackers.
% The result shows low regret for AT-DPT, displaying robustness against this type of attack as well.
The table indicates that for AT-DPT the adaptive attacker is stronger.
AT-DPT (A) performs better than AT-DPT (n-A) when evaluated against an adaptive attacker.
This is unsurprising, but the performance difference is substantial and worth noting.
There is a tradeoff: AT-DPT (n-A) performs better against non-adaptive attackers, though the difference is quite small.
This mirrors the tradeoff observed comparing AT-DPT (n-A) to the baselines -- it outperforms the baselines, but in the clean case it can slightly underperform.
These tradeoffs are not surprising and simply suggest that it is useful to appropriately select the class of attacks expected in the target application.

% Both the attacker and the victim are transformer models, and we study the effect of the number of layers in the transformer on attack performance.
% TODO: do we actually

% ...\\......\\...\\......\\...\\......\\...

% \setlength{\tabcolsep}{-1pt}
% \begin{table*}[b]
%     \footnotesize
%     \centering
%     \begin{tabular}{ccccc|cc}
%         \toprule
%         \multirow{2}{*}{\shortstack[c]{\# of\\layers}} & \multicolumn{4}{c|}{Adv.\ Attacker's Number of Layers} & \multirow[b]{2}{*}{\shortstack[c]{Unif.\ Rand.\\Attack}} & \multirow{2}{*}{Clean Env.} \\
%         & 1 & 2 & 3 & 4 \\
%         \midrule
%         \multicolumn{7}{c}{$\varepsilon = 0.4$} \\
%         \midrule
%         1 & \colorBAdaptiveC{64.9 $\pm$ 17.5}{81.1 $\pm$ 19.9}{69.6 $\pm$ 19.6}{63.9 $\pm$ 23.1}{54.1 $\pm$ 4.2}{68.6 $\pm$ 7.5} \\
%         2 & \colorBAdaptiveC{77.3 $\pm$ 22.7}{70.1 $\pm$ 6.3}{81.7 $\pm$ 15.2}{82.0 $\pm$ 20.0}{60.2 $\pm$ 10.8}{77.1 $\pm$ 9.0} \\
%         3 & \colorBAdaptiveC{64.3 $\pm$ 16.5}{82.1 $\pm$ 15.9}{65.7 $\pm$ 18.8}{67.1 $\pm$ 17.4}{55.8 $\pm$ 6.7}{61.3 $\pm$ 6.2} \\
%         4 & \colorBAdaptiveC{65.8 $\pm$ 21.7}{83.3 $\pm$ 18.5}{61.4 $\pm$ 18.2}{62.6 $\pm$ 14.0}{56.9 $\pm$ 3.5}{64.6 $\pm$ 8.1} \\
%         \bottomrule
%     \end{tabular}
%     \caption{Adaptive attacker bandit results, varying number of layers in the transformer architecture}
%     \label{tab:bandit_adaptive_nlayers}
% \end{table*}
% \setlength{\tabcolsep}{6pt}

\subsection{Linear Bandit Setting}

% !TEX root =  main.tex
%%%%%%%%%%%%%%%%%%%%%%%%%%%%%%%%%%%%%
%%%%%%%%%%%%%%%%%%%%%%%%%%%%%%%%%%%%%

\textbf{Environment.}
We follow a similar setup as in the original DPT work \citep{lee2023dpt}.
We sample $d$-armed linear bandits, where the reward is given by $\E[\,r\mid a, \Mcal_i] = \left\langle \omega_i, \psi(a) \right\rangle$, and $\omega_i \in \R^d$ is a task-specific parameter vector, and $\psi:\Acal \rightarrow \R^d$ is a feature vector shared across all tasks.
Both $\omega_i$ for every $i$ and $\psi$ are sampled from $\Ncal(\mathbf{0}, I_d/d)$.
In our experiments we use $d=2$ and $|\Acal|=10$. %, same as in the original paper.

\textbf{Results.}
From the results in \Cref{tab:linear_bandit_results} we see \hbox{CRLinUCB} performing only marginally better than all other algorithms in the clean case. % and uniform random attack.
Although, under a more complex attack AT-DPT outperforms all other algorithms, and matches \hbox{CRLinUCB} in the clean case. % and uniform random attack.

\setlength{\tabcolsep}{-2pt}
\begin{table*}[b]
    \centering
    \caption{
    Comparison of the cumulative regret (lower is better) of the different algorithms under different attackers in the \textbf{linear bandit setting}, with $\varepsilon=0.4$ steps poisoned. Attack budget $B = 3$. 
    For $\varepsilon=0.1$ and $0.2$ see \Cref{tab:linear_bandit_results_full} in the Appendix.
    Mean and 95\% confidence interval (2$\times$SEM) over 10 replications. 
    $^*$ We use a tuned version of CRLinUCB which outperforms the base version (see \Cref{app:crlinucb}).
    }
    % \vspace{0.5em}
    \begin{footnotesize}
    \begin{tabular}{lcccc|cc}
        \toprule
        \multirow{2}{*}{Algorithm} & \multicolumn{4}{c|}{Attacker Target} & \multirow[b]{2}{*}{\shortstack[c]{Unif.\ Rand.\\Attack}} & \multirow{2}{*}{Clean Env.} \\
        & AT-DPT & DPT & LinUCB & CRLinUCB$^*$ \\
        \midrule
        AT-DPT         & \colorLBC{2.49 $\pm$ 1.06}{2.50 $\pm$ 1.08}{2.83 $\pm$ 1.10}{1.79 $\pm$ 1.02}{5.33 $\pm$ 1.16}{3.89 $\pm$ 0.86} \\
        DPT            & \colorLBC{70.29 $\pm$ 7.32}{71.42 $\pm$ 7.46}{70.83 $\pm$ 7.76}{63.84 $\pm$ 7.18}{6.62 $\pm$ 1.30}{3.35 $\pm$ 0.84} \\
        LinUCB         & \colorLBC{37.69 $\pm$ 4.46}{35.93 $\pm$ 3.86}{35.22 $\pm$ 4.14}{34.82 $\pm$ 4.36}{5.21 $\pm$ 1.16}{3.51 $\pm$ 0.88} \\
        CRLinUCB$^*$   & \colorLBC{37.45 $\pm$ 4.76}{33.03 $\pm$ 4.00}{35.56 $\pm$ 4.26}{35.36 $\pm$ 4.80}{5.12 $\pm$ 1.48}{2.94 $\pm$ 0.78} \\
        \bottomrule
    \end{tabular}
    \end{footnotesize}
    \label{tab:linear_bandit_results}
    \vspace*{-1em}
\end{table*}

\subsection{MDP Setting}

% !TEX root =  main.tex
%%%%%%%%%%%%%%%%%%%%%%%%%%%%%%%%%%%%%
%%%%%%%%%%%%%%%%%%%%%%%%%%%%%%%%%%%%%

\textbf{Environment.}
In the MDP setting we consider an extension of a sparse reward MDP considered in prior work -- the Dark Room environment \citep{lee2023dpt, laskin2022ad, zintgraf2020varibad} -- a 2D gridworld environment where the agent only observes its own state and gains a reward of 1 when at the goal state.
The agent has 5 actions -- $\Acal =\{\text{up}, \text{down}, \text{left}, \text{right}, \text{stay}\}$.
We consider a modification of this -- instead of having one goal, we consider two goals -- one giving a reward of 1, the other giving 2.
To pretrain the DPT we supply optimal actions that lead to the goal giving reward of 2.
We refer to this environment as Darkroom2.

To conform to the sparse reward nature of this environment we constrain the attacker to only output attacks in $\{-1, 0, 1\}$, having a softmax parameterization.
This results in the observed reward being one of $\{-1, 0, 1, 2, 3\}$.
We do not perform any reward normalization or scaling.
In the evaluations we present the underlying episode reward $\sum_h^H \rbar_h$ as the performance metric.
The budget of the attacker $B$ limits the norm of the vector of the attack at every cell.

\textbf{Hyperparameters}
To pretrain DPT in the Darkroom2 setting we use the same model architecture as for the bandit setting, the context length equal to episode length $H=200$, learning rate is $\eta = 0.0001$ and train for $150$ epochs.
For adversarial training we use the learning rate $\eta=0.00003$ for the victim and $\eta_\text{attacker}=0.03$.

\textbf{Evaluation.}
To evaluate AT-DPT we perform cross-validation with different attackers, as in the bandit setting.
For evaluation, we present the total underlying episode reward $\sum_h \rbar_h$ in the tables.
We report the mean and 95\% confidence interval (2$\times$SEM) across 10 different experiment replications.
We run adversarial training for $N=400$ rounds.
The results, seen in \Cref{tab:mdp_megatable_epse0.8}, show that AT-DPT is robust against different attackers, with NPG coming close.
The robustness displayed by NPG has been also observed by \citet{zhang2021robust} -- they find that NPG is robust against $\varepsilon$-contamination if the adversary rewards are bounded.
We also observe that attacks with $\varepsilon = 0.1$ and $\varepsilon = 0.2$ are not very effective for NPG and Q-learning.

In \Cref{tab:miniworld} we additionally show experiments from the Miniworld environment \citep{chevalier2023miniworld}, a 3D environment to evaluate visual navigation from images (25 $\times$ 25 pixels).
We follow a similar setup as in the original DPT paper \citep{lee2023dpt}.
The environment consists of four boxes of different colors, and one of those is chosen as the goal box, unknown to the agent.
The agent receives a reward of $+1$ when stood next to the goal box.
The episode is $H=250$ steps long.
These reseults show, that similarly to the smaller environments AT-DPT is more robust to adversarial reward corruption.
That is, there is evidence to show that the method scales to larger environments.

The main advantage of using AT-DPT over classic algorithms in these scenarios is generalization -- DPT is a meta-learner, which infers the task from a few interactions with the environment and follows an optimal policy almost immediately.
Conversely, NPG and Q-learning are task-specific learners -- they require interactions from the current environment to improve their policies.
In addition, these algorithms require a few (hundreds/thousands of) episodes before converging to a stable policy.
In our experiments we trained different NPG, Q-learning, PPO policies for each environment.
One could argue that it is possible to use a universal policy, but the agent is unaware what the current task is, thus it is unclear what it needs to condition on.

\setlength{\tabcolsep}{-2pt}
\begin{table*}[t]
    \centering
    \caption{Comparison of the average episode reward (higher is better) of the different algorithms under different attackers trained for 300 rounds (5 rounds for Q-learning and NPG) in the \textbf{Darkroom2 environment} (5$\times$5 grid). Mean and 95\% confidence interval (2$\times$SEM) over 10 replications, with $\varepsilon=0.4$ steps poisoned. For $\varepsilon=0.1$ and $0.2$ see \Cref{tab:mdp_megatable_epse0.8_full} in the Appendix.
    Attack budget $B = 10$.
    $^\S$ NPG and Q-learning need multiple episodes of learning to converge to a stable policy; we run them for 100 episodes before evaluating performance.}
    % \vspace{0.5em}
    \footnotesize
    \begin{tabular}{lcccc|cc}
        \toprule
        \multirow{2}{*}{Algorithm} & \multicolumn{4}{c|}{Attacker Target} & \multirow[b]{2}{*}{\shortstack[c]{Unif.\ Rand.\\Attack}} & \multirow{2}{*}{Clean Env.} \\
        & AT-DPT & DPT & NPG & Q-learning \\
        \midrule
        % \multicolumn{7}{c}{$\varepsilon = 0.1$} \\
        % \midrule
        % AT-DPT          & \colorMC{266.4 $\pm$ 18.2}{267.8 $\pm$ 20.7}{262.5 $\pm$ 17.5}{258.9 $\pm$ 20.2}{236.0 $\pm$ 22.5}{273.4 $\pm$ 17.8} \\
        % DPT             & \colorMC{233.6 $\pm$ 8.6}{198.1 $\pm$ 13.0}{220.9 $\pm$ 10.1}{222.6 $\pm$ 6.8}{141.7 $\pm$ 9.9}{305.4 $\pm$ 6.1} \\
        % NPG$^\S$        & \colorMC{241.9 $\pm$ 6.6}{248.1 $\pm$ 6.3}{247.7 $\pm$ 7.1}{243.3 $\pm$ 5.5}{150.8 $\pm$ 2.1}{151.5 $\pm$ 2.5} \\
        % Q-learning$^\S$ & \colorMC{283.5 $\pm$ 3.9}{280.2 $\pm$ 5.1}{246.6 $\pm$ 38.4}{264.3 $\pm$ 17.2}{266.0 $\pm$ 15.7}{263.4 $\pm$ 15.4} \\
        % \midrule
        % \multicolumn{7}{c}{$\varepsilon = 0.2$} \\
        % \midrule
        % AT-DPT          & \colorMC{263.1 $\pm$ 17.6}{269.7 $\pm$ 15.1}{258.0 $\pm$ 17.7}{258.0 $\pm$ 19.3}{245.6 $\pm$ 17.0}{275.8 $\pm$ 16.4} \\
        % DPT             & \colorMC{227.9 $\pm$ 9.5}{169.9 $\pm$ 13.3}{213.6 $\pm$ 6.8}{217.4 $\pm$ 9.5}{131.5 $\pm$ 8.4}{305.4 $\pm$ 6.1} \\
        % NPG$^\S$        & \colorMC{244.1 $\pm$ 7.5}{244.4 $\pm$ 6.7}{239.5 $\pm$ 9.4}{241.2 $\pm$ 9.4}{151.9 $\pm$ 2.1}{151.5 $\pm$ 2.5} \\
        % Q-learning$^\S$ & \colorMC{240.6 $\pm$ 3.6}{254.1 $\pm$ 6.2}{236.5 $\pm$ 6.1}{246.1 $\pm$ 5.3}{246.8 $\pm$ 5.1}{243.3 $\pm$ 6.4} \\
        % \midrule
        % \multicolumn{7}{c}{$\varepsilon = 0.4$} \\
        % \midrule
        AT-DPT     & \colorMC{242.2 $\pm$ 11.9}{267.5 $\pm$ 10.5}{241.7 $\pm$ 10.2}{239.1 $\pm$ 8.8}{258.2 $\pm$ 11.8}{267.4 $\pm$ 15.1} \\
        AT-DPT (sub.\ 10\%)\ \ & \colorMC{225.2 $\pm$ 17.8}{245.5 $\pm$ 17.9}{226.1 $\pm$ 16.6}{222.7 $\pm$ 17.6}{241.3 $\pm$ 17.8}{250.2 $\pm$ 18.8} \\
        DPT        & \colorMC{216.1 $\pm$ 11.0}{143.5 $\pm$ 11.0}{202.6 $\pm$ 7.4}{205.9 $\pm$ 7.8}{266.2 $\pm$ 8.1}{306.8 $\pm$ 7.1} \\
        NPG$^\S$ & \colorMC{237.2 $\pm$ 6.7}{243.7 $\pm$ 7.9}{228.9 $\pm$ 4.0}{228.1 $\pm$ 8.1}{235.3 $\pm$ 8.2}{241.7 $\pm$ 7.5} \\
        Q-learning$^\S$ & \colorMC{198.1 $\pm$ 3.7}{238.6 $\pm$ 6.0}{215.4 $\pm$ 7.6}{224.7 $\pm$ 7.3}{229.0 $\pm$ 7.2}{225.6 $\pm$ 5.4} \\
        \bottomrule
    \end{tabular}
    \label{tab:mdp_megatable_epse0.8}
    \vspace*{-0.5em}
\end{table*}

\setlength{\tabcolsep}{-2.5pt}
\begin{table}[b]
    \footnotesize
    \centering
    \caption{
        Comparison of avg.\ episode reward (higher is better) of the algorithms under different attackers trained for 100 rounds in the \textbf{Miniworld environment} with $\varepsilon = 0.4$ steps poisoned.
        Mean and 95\% confidence interval (2$\times$SEM) over 10 runs.
        Attack budget $B = 5$.
        For $\varepsilon = 0.1$ and $0.2$ see \Cref{tab:miniworld_full} in the Appendix.
        $^\S$ PPO needs multiple episodes of online learning to converge to a stable policy; we run it for 100 episodes before evaluating performance.
    }
    \begin{adjustbox}{max width=\linewidth}
    \begin{tabular}{lcc|cc}
        \toprule
        \multirow{2}{*}{Algorithm\!\!\!} & \multicolumn{2}{c|}{Attacker Target} & \multirow[b]{2}{*}{\shortstack[c]{Unif.\ Rand.\\Attack}} & \multirow{2}{*}{Clean Env.} \\
        & AT-DPT & DPT \\
        \midrule
        AT-DPT\!\!      & \colorMw{104.8 $\pm$ 16.0}{116.8 $\pm$ 18.8}{108.6 $\pm$ 15.1}{112.7 $\pm$ 23.9} \\
        DPT             & \colorMw{81.2 $\pm$ 12.2}{70.2 $\pm$ 15.0}{102.7 $\pm$ 13.1}{110.0 $\pm$ 14.7} \\
        PPO$^\S$        & \colorMw{83.5 $\pm$ 7.4}{83.8 $\pm$ 7.2}{92.9 $\pm$ 7.3}{123.5 $\pm$ 8.1} \\
        \bottomrule
    \end{tabular}
    \end{adjustbox}
    \label{tab:miniworld}
\end{table}

\setlength{\tabcolsep}{6pt}

% !TEX root =  main.tex
%%%%%%%%%%%%%%%%%%%%%%%%%%%%%%%%%%%%%
%%%%%%%%%%%%%%%%%%%%%%%%%%%%%%%%%%%%%
\section{Discussion}\label{sec:discussion}

\textbf{Significance.}
The key contribution of this work is the setup itself -- a set of provided ingredients which lead to test-time adversarial robustness, which has not been considered in prior work.
This is done via simultaneously training a population of attackers minimizing the underlying environment rewards, and the victim optimizing for the optimal actions from the poisoned data.
By showing extensive evaluations in the bandit and MDP setting we demonstrated that adversarial training can work for in-context RL, showing generalization across different attacks (incl.\ adaptive).

Our approach significantly outperforms baselines that do not account for the multi-task structure (e.g., \Cref{tab:linear_bandit_results}, by more than 90\%).
We believe there are two reasons.
First, the algorithm can use an informed prior over the distribution of tasks, making corruption schemes that are implausible according to the prior ineffective.
Second, the algorithm can be thought of as posterior sampling that uses ``corrected'' posterior, which accounts for the corruption process.
% \textcolor{red}{In addition many papers only consider robustness to adversarial attacks which are in the same distribution as in training \citep{zhang2021robust}. That means these methods are only robust to the specific attack they are run against, but in our case AT-DPT is robust against attacks that differ from its training.}

% \textcolor{blue}{We are not aware of works prior to ours that considered poisoning attacks for in-context RL.}

The method introduced in this work is based on DPT \citep{lee2023dpt}, although extensions to other ICRL works could be possible, such as \textit{Vintix} \citep{polubarov2025vintix}, or \textit{ICPE} \citep{russo2026incontext}, which we leave for future work.

\textbf{Theoretical Insights.}
At a high level, our approach can be viewed as approximately solving the bi-level \mbox{objective}:
\( \min_{\theta} \E_{\Mcal_i, D \sim \Mcal_i(\pi_{\theta}, \pi_{\phi, i}),s_{\text{q}}} \left[ \ell(\pi_{\theta}(\cdot | s_{\text{q}}, D), a^*) \right] \) such that \(\cramped{\pi_{\phi, i} \in \mathrm{BR}(\Mcal_i, \pi_{\theta})}\) for all tasks $\Mcal_i$,
where $\mathrm{BR}$ denotes the set of best responses of the adversary given the agent's policy, and $\cramped{D \sim \Mcal_i(\pi_{\theta},\pi_{\phi,i})}$ is collected under attacker $\pi_{\phi,i}$ and agent $\pi_\theta$, as specified in \Cref{sec:attack_model}.
AT-DPT's adversaries are trained to minimize the agent's return under a soft budget constraint, therefore the resulting policy $\pi_{\theta}$ is expected to perform robustly against a range of adversarial strategies.
This is consistent with empirical results -- we observe that AT-DPT's performance does not substantially degrade when evaluated against unseen adversaries.
In equilibrium $(\theta^*, \phi^*)$ we have
$
{\theta^* \in \argmin_{\theta} \E_{\Mcal_i, D \sim \Mcal_i(\pi_{\theta}, \pi_{\phi^*, i}), s_{\text{q}}} \left[ \ell (\pi_{\theta}( \cdot | s_{\text{q}}, D), a^*) \right]}
$.
Theorem 1 of \citet{lee2023dpt} suggests that $\pi_{\theta^*}$ implements in-context posterior sampling, but with a \textit{corrected} posterior that accounts for adversarial data corruption (i.e., $\pi_{\phi^*}$).
As long as the underlying reward means remain identifiable under the corruption process, standard posterior sampling analyses are applicable.
In this case, the regret of a learning algorithm implemented by $\pi_{\theta^*}$ depends on the extent of corruption since increased corruption that reduces reward separability leads to higher regret \citep[by analogy with the lower bound by][]{lai1985bandits}.
Given that $\pi_{\phi^*}$ minimizes the agent's return (under a constraint) for $\pi_{\theta^*}$, the guarantees on $\pi_{\theta^*}$ hold when the test-time corruption differs from $\pi_{\phi^*}$ but satisfies the budget constraint.
A complete theoretical account would require establishing that AT-DPT converges to an approximate equilibrium of the underlying game, and that the approximate nature of the resulting solution does not significantly affect the results.
We believe that performing such an analysis is challenging and therefore leave it to future work.

\textbf{Practical Relevance.}
In general, robustness to test-time poisoning attacks is important for agents relying on in-context learning.
A prime example are LLM-based agents, whose context can be easily poisoned if they use external tools or RAG mechanisms, as demonstrated by prior work \citep{greshake2023not,carlini2024poisoning}.
Our work is generally important for understanding how to enable robustness to poisoning attacks for agents that exhibit ICRL capabilities -- as shown by \citet{monea2024llms}, this includes LLMs.
One of our goals was to provide a principled approach to learning a robust RL algorithm which better utilizes prior knowledge, which we can tailor to specific tasks of interest (if our prior is informed).
This may yield significant performance gains, and our empirical results support this (e.g., see \Cref{tab:linear_bandit_results}).

\textbf{Limitations and Future Work.}
The main limitation of our method, also a limitation of DPT is the need of actions provided by the oracle for training \citep{lee2023dpt}.
The authors of DPT propose relaxing this requirement by supplying actions generated by another RL agent which performs sufficiently well for the current task, but this might not be possible in an adversarial scenario.
A different approach is viable -- training on offline trajectories with simulated attackers.
We provide experiments demonstrating the performance of AT-DPT if suboptimal demonstrations were given (\textit{subopt.}\ rows in \Cref{tab:bandit_megatable_epse0.8,tab:miniworld}) -- AT-DPT allows for some percentage of suboptimal demonstrations, but the performance degrades when increasing this percentage.

We also observe in our results the capability of AT-DPT to generalize beyond the attack it has been trained on (i.e., adversarially trained against its own specific attacker, generalizes to an attacker trained for TS, for example).
This suggests it may be possible to exploit this further by adversarially training AT-DPT with multiple different contamination levels $\varepsilon$.
Additionally in our results we only consider a single attack specification per experiment.
To make AT-DPT even more robust, and potentially alleviate the trade-off observed in the clean and random attack environment it would be possible to train AT-DPT with multiple different attack specifications (e.g., mixing in non-adaptive and adaptive attacks), which we leave as a direction for future work.

\section*{Code Availability}
Code available at \href{https://github.com/PauliusSasnauskas/AT-DPT}{github.com/PauliusSasnauskas/AT-DPT}.

% % Acknowledgements should only appear in the accepted version.
\section*{Acknowledgements}
This research was, in part, funded by the Deutsche Forschungsgemeinschaft (DFG, German Research Foundation) -- project number 467367360.
We acknowledge the use of computing resources generously provided by \mbox{MPI-SWS}.
% If a paper is accepted, the final camera-ready version can (and usually should)
% include acknowledgements.  Such acknowledgements should be placed at the end of
% the section, in an unnumbered section that does not count towards the paper
% page limit. Typically, this will include thanks to reviewers who gave useful
% comments, to colleagues who contributed to the ideas, and to funding agencies
% and corporate sponsors that provided financial support.

\section*{Impact Statement}
Adversarial data poisoning attacks have potential to be used by malicious actors to generate attacks against real systems.
Therefore, as is highlighted in the paper, it is important to study methods that are robust.
Our work presents a strong case for robustifying against these kinds of attacks -- it can enable us to design more efficient algorithms that account for the distribution of tasks and utilize this information as a prior to obtain better robustness.
Data poisoning attacks have existed in the literature before, and have many potential societal consequences, none which we feel must be specifically highlighted here.

\bibliography{main}
\bibliographystyle{icml2026}

%%%%%%%%%%%%%%%%%%%%%%%%%%%%%%%%%%%%%%%%%%%%%%%%%%%%%%%%%%%%%%%%%%%%%%%%%%%%%%%
%%%%%%%%%%%%%%%%%%%%%%%%%%%%%%%%%%%%%%%%%%%%%%%%%%%%%%%%%%%%%%%%%%%%%%%%%%%%%%%
% APPENDIX
%%%%%%%%%%%%%%%%%%%%%%%%%%%%%%%%%%%%%%%%%%%%%%%%%%%%%%%%%%%%%%%%%%%%%%%%%%%%%%%
%%%%%%%%%%%%%%%%%%%%%%%%%%%%%%%%%%%%%%%%%%%%%%%%%%%%%%%%%%%%%%%%%%%%%%%%%%%%%%%
% \newpage
\clearpage
\appendix
\onecolumn
% !TEX root =  main.tex
%%%%%%%%%%%%%%%%%%%%%%%%%%%%%%%%%%%%%
%%%%%%%%%%%%%%%%%%%%%%%%%%%%%%%%%%%%%

% Restore addcontentsline functionality broken by icml2026.sty
\makeatletter
\pretocmd{\@startsection}{\phantomsection}{}{}
\makeatother

\vspace*{-1em}
{\centering {\LARGE\bf Supplementary Material:\papertitle\par}}

\hypersetup{linkcolor=black}  % set link color to black for the table of contents
\renewcommand{\contentsname}{\vspace*{-1em}}
\addtocontents{toc}{\protect\setcounter{tocdepth}{3}}  % Show up to subsections
\tableofcontents
\clearpage
\renewcommand{\floatpagefraction}{.95}

\hypersetup{linkcolor=mydarkblue}

\section{Baseline algorithms}\label{app:baselines}

We believe a comparison with Algorithm Distillation \citep[AD,][]{laskin2022ad} would be unfair to AD, because in contrast to DPT, AD is learning to imitate a learning algorithm.
The two methods are conceptually different: AD distills an RL algorithm learning how to learn (i.e., predict an action by implicitly performing a policy improvement, given the previous steps), whereas DPT instead is trained on optimal actions (i.e., predict optimal action by implicitly understanding the goal, similarly to posterior sampling, given the previous steps).
Furthermore, following the original setup of the AD paper it would be quite hard to collect training data given our attack model -- there is no straightforward way to get trajectories displaying a policy improvement under an attacker.
Regardless, we include a few experiments with AD in \Cref{tab:bandit_megatable_epse0.8_appendix} for $\varepsilon = 0.4$.

\subsection{Robust TS}\label{app:rts}
\citet{xu2024robustts} provide the Robust TS algorithm.
This algorithm relies on a corruption level hyperparameter $\Cbar$.
The recommendation given by the authors is to set this to $\sum_h^H c_h \leq \Cbar$, where $c_h$ is the corruption level (i.e., in our case, $c_h = r_h - \rbar_h$) for step $h$, if the corruption level is known.
If the corruption level is unknown, the authors suggest setting $\Cbar = \sqrt{H \frac{\ln |\Acal|}{|\Acal|}}$.

Following these recommendations, in an environment with $H = 500$, $|\Acal| = 5$, $\varepsilon = 0.4$, our preliminary findings are:
\begin{itemize}
    \item assume corruption level is known: $\Cbar \approx 120$ -- RTS performance is worse than TS; indicated as RTS ($\Cbar$ known);
    \item assume corruption level is unknown: $\Cbar \approx 12.7$ -- RTS performance is worse than TS; indicated as RTS ($\Cbar$ unk.);
    \item tuned $\Cbar$ for our setup: $\Cbar = 0.5$ -- RTS performance is better than TS; indicated as RTS ($\Cbar$ tuned).
\end{itemize}
We report the best scores (obtained with the tuned variant) in the main text, giving the full three variant comparison in \Cref{tab:bandit_megatable_epse0.8_appendix}.

\subsection{crUCB}\label{app:crucb}

\citet{niss2020crucb} provide a few variants of the crUCB algorithm.
We chose the $\alpha$-trimmed variant, which performs best empirically.
We introduce a modification to the algorithm due to poor original variant empirical performance.
The modified variant is shown in \cref{alg:crucb-mod}, where $f$ -- $\alpha$-trimmed mean function -- if $n$ is the number of rewards observed for that arm, removes $\left\lceil \alpha\, n \right\rceil$ lowest and $\left\lceil \alpha\, n \right\rceil$ highest rewards observed for that specific arm; removing $2 \left\lceil \alpha\, n \right\rceil$ elements in total, and $\mathbf{x}_a^{(h)}$ -- list of observed rewards for arm $a$ at step $h$.

\begin{algorithm}[htb]
\caption{crUCB ($\alpha$-trimmed variant), modified}
\label{alg:crucb-mod}
\begin{algorithmic}[1]
    \STATE \textbf{input:} $\alpha$ -- fraction of steps poisoned
    \STATE \textbf{input:} $\sigma_0$ -- upper bound on sub-Gaussian constant (hyperparameter)
    \STATE \textbf{input:} $f$ -- mean estimate function ($\alpha$-trimmed mean)
    \FOR{step $h = 1, \dots, H$}
        \FOR{each $a \in \Acal$}
            \STATE $\hat{\mu}^{(h)}_a \gets f(\mathbf{x}^{(h)}_a)$ \quad ($\alpha$-trimmed mean estimate of rewards)
            \STATE $N_a^{(h)} \gets$ number of times action $a$ has been played
        \ENDFOR
        \STATE Choose action $a = \argmax_{a \in \Acal} \hat{\mu}^{(h)}_a + {\color{blue} \sigma_0} \left( \sqrt{\frac{4 \log(h)}{{\color{blue} \left\lfloor (1 - 2\alpha)\, N^{(h)}_a \right\rfloor}}} \right)$
    \ENDFOR
\end{algorithmic}
\end{algorithm}

The original bonus term in the algorithm is $\displaystyle {\color{red} \frac{\sigma_0}{1 - 2 \alpha}} \left( \sqrt{\frac{4 \log(h)}{{\color{red} N^{(h)}_a}}} \right)$.

Assume $f(\mathbf{z})$ with $n$ elements returns zero if $\mathbf{z}$ contains fewer than $n - 2 \left\lceil \alpha\, n \right\rceil$ elements.
The failure is observed when the assumption above is true -- the estimated mean returns zero, whereas the bonus is not infinity, leading to arms which have only been played one time have a very low score.

We report the best scores (obtained with the modified variant) in the main text, giving full results in \Cref{tab:bandit_megatable_epse0.8_appendix} comparing:
\begin{itemize}
    \item the original variant, indicated as crUCB (orig.) or (o.);
    \item the original variant with $\sigma_0$ scaled by $\sqrt{1 - 2 \alpha}$, indicated as crUCB (low $\sigma_0$) or (l.\ $\sigma_0$);
    \item the modified variant, indicated as crUCB (mod.) or (m.).
\end{itemize}

\subsection{CRLinUCB}\label{app:crlinucb}
We source the CRLinUCB algorithm from \citet{ding2022crlinucb}.
The authors suggest setting the upper bound of the budget $C'$ to equal $\varepsilon B H$.
We found that the algorithm did not perform well when set to this value.
We then tuned this variant, and present a number of results in the tables:
\begin{itemize}
    \item the original variant, denoted as CRLinUCBv1, where the hyperparameters are set to the values suggested by Theorem 1 by \citet{ding2022crlinucb};
    \item the variant where the bound is divided by the time horizon $H$, denoted as CRLinUCBv2, which approximately matches the values of the experiments of \citet{ding2022crlinucb};
    \item a third variant, CRLinUCBv3, where the hyperparameters are interpolated between v1 and v2, they are within the same order of magnitude with the geometric mean of the values used in v1 and v2.
\end{itemize}
In the main text we report the results from CRLinUCBv2, which seemed to work best in our case.

\newcommand{\colorABA}[8]{%
  \colorizeValue[18mm][0.8mm]{10}{60}{#1} &
  \colorizeValue[18mm][0.8mm]{10}{60}{#2} &
  \colorizeValue[18mm][0.8mm]{10}{60}{#3} &
  \colorizeValue[18mm][0.8mm]{10}{60}{#4} &
  \colorizeValue[18mm][0.8mm]{10}{60}{#5} &
  \colorizeValue[18mm][0.8mm]{10}{60}{#6} &
  \colorizeValue[18mm][0.8mm]{10}{60}{#7} &
  \colorizeValue[18mm][0.8mm]{10}{60}{#8}%
}
\newcommand{\colorABAclean}[2]{%
  \colorizeValue[18mm][0.8mm]{10}{60}{#1} &
  \colorizeValue[18mm][0.8mm]{10}{60}{#2}%
}

\newcommand{\colorABB}[8]{%
  \colorizeValue[18mm][0.8mm]{20}{90}{#1} &
  \colorizeValue[18mm][0.8mm]{20}{90}{#2} &
  \colorizeValue[18mm][0.8mm]{20}{90}{#3} &
  \colorizeValue[18mm][0.8mm]{20}{90}{#4} &
  \colorizeValue[18mm][0.8mm]{20}{90}{#5} &
  \colorizeValue[18mm][0.8mm]{20}{90}{#6} &
  \colorizeValue[18mm][0.8mm]{20}{90}{#7} &
  \colorizeValue[18mm][0.8mm]{20}{90}{#8}%
}
\newcommand{\colorABBclean}[2]{%
  \colorizeValue[18mm][0.8mm]{10}{70}{#1} &
  \colorizeValue[18mm][0.8mm]{10}{70}{#2}%
}

\newcommand{\colorABC}[8]{%
  \colorizeValue[18mm][0.8mm]{40}{110}{#1} &
  \colorizeValue[18mm][0.8mm]{40}{110}{#2} &
  \colorizeValue[18mm][0.8mm]{40}{110}{#3} &
  \colorizeValue[18mm][0.8mm]{40}{110}{#4} &
  \colorizeValue[18mm][0.8mm]{40}{110}{#5} &
  \colorizeValue[18mm][0.8mm]{40}{110}{#6} &
  \colorizeValue[18mm][0.8mm]{40}{110}{#7} &
  \colorizeValue[18mm][0.8mm]{40}{110}{#8}%
}
\newcommand{\colorABCclean}[2]{%
  \colorizeValue[18mm][0.8mm]{10}{70}{#1} &
  \colorizeValue[18mm][0.8mm]{10}{70}{#2}%
}
\newcommand{\colorABbudget}[8]{%
  \colorizeValue[18mm][0.8mm]{20}{100}{#1} &
  \colorizeValue[18mm][0.8mm]{20}{100}{#2} &
  \colorizeValue[18mm][0.8mm]{20}{100}{#3} &
  \colorizeValue[18mm][0.8mm]{20}{100}{#4} &
  \colorizeValue[18mm][0.8mm]{20}{100}{#5} &
  \colorizeValue[18mm][0.8mm]{20}{100}{#6} &
  \colorizeValue[18mm][0.8mm]{20}{100}{#7} &
  \colorizeValue[18mm][0.8mm]{20}{100}{#8}%
}

\section{Additional results}
\subsection{Bandit setting}\label{app:bandit_setting}

We present \Cref{tab:bandit_megatable_epse0.8_appendix}, which shows the full set of results from the bandit setting.
Note, that RTS ($\bar{C}$ unk.), RTS ($\bar{C}$ known) did not perform well, as also noted in \Cref{app:rts}.
Similarly, crUCB (orig.) did not perform well, as noted in \Cref{app:crucb}.
The high values obtained in the case, where the attacker is crUCB (o.) mean that the attacker trained against this algorithm was not performing well, and therefore led to a weak attack.
Recall that the setup has a dual objective, and simply judging by the regret or reward of a single row or column is not enough.

We additionally include AT-DPT trained against a uniform random attack (indicated by AT-DPT (rand.)).
The results for AT-DPT (rand.) show that training against random noise improves robustness in many cases, but is substantially worse than training against a strong adversary.
This means for better robustness it is important to train a strong adversary.

We include experiments with Algorithm Distillation \citep[AD,][]{laskin2022ad} for the case when $\varepsilon = 0.4$.
We train AD with source algorithms TS (row \textit{AD [TS]}), crUCB (m.) (row \textit{AD [crUCB (m.)]}), and crUCB (m.) under corruption (row \textit{AD [crUCB (m.) c.]}).
All instances of AD were trained with embedding dimension 32, 4 layers, 4 attention heads, 0 dropout, 1e-4 learning rate, AdamW (with 1e-4 weight decay), batch size 64, but the experiment setup is otherwise the same as in the rest of the table.
It seems that AD with a sufficiently good source algorithm (e.g., AD with source crUCB) does perform better than the source algorithm alone (e.g., just crUCB), which performs slightly better than simply DPT alone.
Out of all of these AT-DPT still seems to have the lowest regret, which is expected, because it is trained to be robust to corruption.
As mentioned previously, to train robustly for AD we would have to show policy improvement under corruption, which is difficult on its own; alternatively, supplying optimal actions to AD during adversarial training would just make it a DPT method.

\clearpage

\setlength{\tabcolsep}{-2pt}
\begin{table*}[!htb]
    \begin{footnotesize}
    \centering
    \caption{Comparison of the cumulative regret (lower is better) of the different algorithms under different attackers trained for 20 rounds, in the bandit setting. Mean and 95\% confidence interval (2$\times$SEM) over 10 experiment replications. Attack budget $B = 3$.}
    \label{tab:bandit_megatable_epse0.8_appendix}
    \hspace*{-3.5em}
    \begin{adjustbox}{bgcolor=white,margin=0 0 0 -0.3em}
    \begin{tabular}{lcccccccc|cc}
        \toprule
        \multirow{2}{*}{Algorithm} & \multicolumn{8}{c|}{Attacker Target} & \multirow[b]{2}{*}{\shortstack[c]{Unif.\ Rand.\\Attack}} & \multirow{2}{*}{Clean Env.} \\
        & AT-DPT & DPT & TS & RTS ($\Cbar$ t.) & UCB1.0 & crUCB (o.) & crUCB (l.\ $\sigma_0$) & crUCB (m.) \\
        \midrule
        \multicolumn{11}{c}{$\varepsilon = 0.1$} \\
        \midrule
        AT-DPT                & \colorABA{14.5 $\pm$ 0.9}{13.9 $\pm$ 0.7}{14.4 $\pm$ 0.8}{14.6 $\pm$ 1.0}{14.8 $\pm$ 0.8}{14.4 $\pm$ 1.4}{14.1 $\pm$ 0.6}{14.2 $\pm$ 1.3} & \colorABAclean{14.2 $\pm$ 1.2}{12.4 $\pm$ 1.1} \\
        AT-DPT (sub.\ 10\%)	  & \colorABA{21.0 $\pm$ 1.6}{20.9 $\pm$ 1.7}{21.0 $\pm$ 1.4}{21.1 $\pm$ 1.4}{21.1 $\pm$ 1.6}{20.8 $\pm$ 1.6}{20.6 $\pm$ 1.6}{20.9 $\pm$ 1.9} & \colorABAclean{19.5 $\pm$ 2.0}{17.2 $\pm$ 1.8} \\
        AT-DPT (sub.\ 20\%)	  & \colorABA{26.1 $\pm$ 1.7}{26.0 $\pm$ 1.9}{26.0 $\pm$ 1.9}{26.0 $\pm$ 1.7}{26.3 $\pm$ 1.5}{26.0 $\pm$ 1.7}{25.6 $\pm$ 1.8}{25.6 $\pm$ 2.0} & \colorABAclean{24.0 $\pm$ 2.1}{21.9 $\pm$ 2.2} \\
        AT-DPT (sub.\ 30\%)	  & \colorABA{32.1 $\pm$ 2.3}{31.9 $\pm$ 2.4}{32.0 $\pm$ 2.2}{32.2 $\pm$ 2.2}{32.2 $\pm$ 2.2}{31.8 $\pm$ 2.0}{31.7 $\pm$ 2.3}{31.8 $\pm$ 2.4} & \colorABAclean{29.1 $\pm$ 2.3}{27.0 $\pm$ 2.5} \\
        AT-DPT (rand.)        & \colorABA{18.4 $\pm$ 1.2}{17.9 $\pm$ 1.0}{17.4 $\pm$ 1.0}{18.1 $\pm$ 1.6}{17.9 $\pm$ 0.9}{18.3 $\pm$ 1.5}{17.4 $\pm$ 1.2}{17.9 $\pm$ 1.6} & \colorABAclean{14.6 $\pm$ 1.1}{12.9 $\pm$ 1.1} \\
        DPT                   & \colorABA{24.2 $\pm$ 1.8}{22.7 $\pm$ 2.0}{22.1 $\pm$ 1.3}{23.0 $\pm$ 1.6}{22.2 $\pm$ 1.6}{21.2 $\pm$ 1.8}{22.6 $\pm$ 1.5}{22.2 $\pm$ 1.8} & \colorABAclean{15.2 $\pm$ 0.8}{12.1 $\pm$ 0.5} \\
        TS                    & \colorABA{27.9 $\pm$ 1.1}{26.6 $\pm$ 2.1}{28.4 $\pm$ 1.8}{27.2 $\pm$ 1.6}{25.9 $\pm$ 1.6}{22.4 $\pm$ 2.1}{28.0 $\pm$ 1.8}{28.6 $\pm$ 1.8} & \colorABAclean{12.0 $\pm$ 1.0}{8.9 $\pm$ 0.4} \\
        RTS ($\Cbar$ tuned)   & \colorABA{27.1 $\pm$ 0.6}{27.0 $\pm$ 1.4}{26.9 $\pm$ 1.0}{26.5 $\pm$ 2.0}{24.2 $\pm$ 1.1}{21.3 $\pm$ 1.4}{26.2 $\pm$ 1.4}{26.8 $\pm$ 1.2} & \colorABAclean{13.0 $\pm$ 0.8}{10.5 $\pm$ 0.3} \\
        RTS ($\Cbar$ unk.)    & \colorABA{59.8 $\pm$ 0.5}{59.4 $\pm$ 0.8}{59.2 $\pm$ 0.6}{59.7 $\pm$ 0.7}{58.7 $\pm$ 0.6}{55.7 $\pm$ 1.0}{59.2 $\pm$ 1.0}{59.1 $\pm$ 0.7} & \colorABAclean{49.9 $\pm$ 0.8}{49.3 $\pm$ 0.8} \\
        RTS ($\Cbar$ known)   & \colorABA{94.9 $\pm$ 0.8}{94.4 $\pm$ 0.9}{94.5 $\pm$ 0.8}{94.6 $\pm$ 1.2}{94.0 $\pm$ 0.8}{91.8 $\pm$ 1.1}{93.8 $\pm$ 1.1}{94.4 $\pm$ 1.0} & \colorABAclean{84.7 $\pm$ 1.6}{84.1 $\pm$ 1.7} \\
        UCB1.0                & \colorABA{30.8 $\pm$ 1.5}{28.5 $\pm$ 1.5}{29.5 $\pm$ 0.8}{28.5 $\pm$ 1.8}{27.3 $\pm$ 1.5}{24.5 $\pm$ 1.0}{28.7 $\pm$ 1.5}{29.4 $\pm$ 1.3} & \colorABAclean{17.9 $\pm$ 0.4}{16.1 $\pm$ 0.3} \\
        crUCB (orig.)         & \colorABA{82.4 $\pm$ 0.8}{81.9 $\pm$ 0.7}{82.5 $\pm$ 0.9}{81.9 $\pm$ 1.3}{82.1 $\pm$ 0.7}{81.3 $\pm$ 1.0}{82.1 $\pm$ 0.8}{81.9 $\pm$ 0.8} & \colorABAclean{79.2 $\pm$ 1.4}{79.3 $\pm$ 1.5} \\
        crUCB (low $\sigma_0$)& \colorABA{19.5 $\pm$ 1.8}{19.1 $\pm$ 1.2}{20.0 $\pm$ 1.0}{18.8 $\pm$ 2.1}{20.1 $\pm$ 1.5}{18.6 $\pm$ 1.7}{19.9 $\pm$ 1.6}{19.6 $\pm$ 2.0} & \colorABAclean{11.1 $\pm$ 0.7}{9.3 $\pm$ 0.5} \\
        crUCB (mod.)          & \colorABA{19.4 $\pm$ 1.7}{18.4 $\pm$ 1.2}{20.5 $\pm$ 1.2}{17.8 $\pm$ 1.6}{19.7 $\pm$ 1.1}{18.7 $\pm$ 1.6}{19.5 $\pm$ 1.0}{18.4 $\pm$ 1.6} & \colorABAclean{11.0 $\pm$ 0.7}{9.2 $\pm$ 0.3} \\
        \midrule
        \multicolumn{11}{c}{$\varepsilon = 0.2$} \\
        \midrule
        AT-DPT                & \colorABB{17.9 $\pm$ 1.4}{17.1 $\pm$ 1.4}{19.0 $\pm$ 1.5}{18.4 $\pm$ 1.3}{17.8 $\pm$ 1.4}{17.4 $\pm$ 1.9}{17.0 $\pm$ 0.9}{16.9 $\pm$ 1.2} & \colorABBclean{20.4 $\pm$ 2.0}{14.5 $\pm$ 1.8} \\
        AT-DPT (sub.\ 10\%)   & \colorABB{24.7 $\pm$ 1.3}{23.7 $\pm$ 1.5}{25.9 $\pm$ 1.9}{24.3 $\pm$ 1.9}{24.0 $\pm$ 1.5}{23.5 $\pm$ 1.6}{23.6 $\pm$ 1.2}{23.4 $\pm$ 1.6} & \colorABBclean{24.7 $\pm$ 1.7}{17.3 $\pm$ 1.7} \\
        AT-DPT (sub.\ 20\%)   & \colorABB{29.7 $\pm$ 1.8}{30.1 $\pm$ 2.0}{31.2 $\pm$ 2.2}{30.4 $\pm$ 2.1}{30.5 $\pm$ 1.9}{29.3 $\pm$ 1.7}{29.4 $\pm$ 1.8}{29.3 $\pm$ 1.8} & \colorABBclean{28.9 $\pm$ 2.0}{22.0 $\pm$ 2.1} \\
        AT-DPT (sub.\ 30\%)   & \colorABB{35.6 $\pm$ 2.1}{35.8 $\pm$ 2.4}{36.8 $\pm$ 2.2}{36.2 $\pm$ 2.2}{35.5 $\pm$ 2.4}{34.6 $\pm$ 1.9}{35.0 $\pm$ 2.3}{35.0 $\pm$ 2.1} & \colorABBclean{33.8 $\pm$ 2.1}{27.0 $\pm$ 2.4} \\
        AT-DPT (rand.)        & \colorABB{23.9 $\pm$ 2.1}{24.3 $\pm$ 2.0}{24.8 $\pm$ 2.5}{24.5 $\pm$ 2.4}{24.1 $\pm$ 2.3}{23.9 $\pm$ 2.3}{23.6 $\pm$ 2.5}{24.1 $\pm$ 2.0} & \colorABBclean{21.0 $\pm$ 2.0}{15.5 $\pm$ 2.0} \\
        DPT                   & \colorABB{35.2 $\pm$ 4.1}{33.2 $\pm$ 3.5}{37.1 $\pm$ 5.2}{35.1 $\pm$ 3.1}{33.0 $\pm$ 3.5}{28.9 $\pm$ 3.1}{35.1 $\pm$ 3.7}{33.1 $\pm$ 3.9} & \colorABBclean{22.3 $\pm$ 1.1}{12.1 $\pm$ 0.5} \\
        TS                    & \colorABB{51.1 $\pm$ 3.2}{48.6 $\pm$ 3.5}{51.1 $\pm$ 3.1}{50.1 $\pm$ 2.7}{47.2 $\pm$ 2.4}{33.7 $\pm$ 3.3}{51.8 $\pm$ 1.7}{49.9 $\pm$ 3.8} & \colorABBclean{18.7 $\pm$ 1.1}{8.9 $\pm$ 0.4} \\
        RTS ($\Cbar$ tuned)   & \colorABB{49.8 $\pm$ 3.4}{44.8 $\pm$ 2.7}{48.1 $\pm$ 2.0}{48.3 $\pm$ 3.1}{44.3 $\pm$ 3.8}{32.4 $\pm$ 2.5}{47.6 $\pm$ 2.0}{46.3 $\pm$ 1.7} & \colorABBclean{19.1 $\pm$ 1.1}{10.5 $\pm$ 0.3} \\
        RTS ($\Cbar$ unk.)    & \colorABB{76.0 $\pm$ 2.1}{73.2 $\pm$ 1.2}{74.5 $\pm$ 1.7}{74.2 $\pm$ 1.7}{72.1 $\pm$ 1.3}{63.5 $\pm$ 1.2}{73.8 $\pm$ 1.2}{73.7 $\pm$ 1.2} & \colorABBclean{53.0 $\pm$ 0.9}{49.3 $\pm$ 0.8} \\
        RTS ($\Cbar$ known)   & \colorABB{131.7 $\pm$ 1.8}{130.7 $\pm$ 1.5}{131.3 $\pm$ 1.5}{130.9 $\pm$ 1.5}{130.6 $\pm$ 1.5}{127.0 $\pm$ 1.6}{130.4 $\pm$ 1.5}{131.0 $\pm$ 1.6} & \colorABBclean{116.4 $\pm$ 2.5}{115.7 $\pm$ 2.6} \\
        UCB1.0                & \colorABB{51.9 $\pm$ 2.5}{46.7 $\pm$ 1.8}{50.8 $\pm$ 1.9}{47.3 $\pm$ 1.9}{45.2 $\pm$ 2.7}{34.2 $\pm$ 2.3}{49.1 $\pm$ 1.8}{48.1 $\pm$ 2.7} & \colorABBclean{23.9 $\pm$ 0.8}{16.1 $\pm$ 0.3} \\
        crUCB (orig.)         & \colorABB{101.6 $\pm$ 1.2}{100.8 $\pm$ 1.2}{101.7 $\pm$ 1.3}{100.5 $\pm$ 1.6}{101.0 $\pm$ 1.0}{98.8 $\pm$ 1.3}{101.1 $\pm$ 1.1}{101.3 $\pm$ 1.2} & \colorABBclean{96.0 $\pm$ 2.0}{95.6 $\pm$ 2.0} \\
        crUCB (low $\sigma_0$)& \colorABB{34.7 $\pm$ 2.1}{33.6 $\pm$ 2.1}{34.4 $\pm$ 2.1}{31.6 $\pm$ 2.8}{32.9 $\pm$ 1.6}{30.5 $\pm$ 3.0}{32.9 $\pm$ 2.2}{34.1 $\pm$ 1.9} & \colorABBclean{15.2 $\pm$ 0.6}{9.4 $\pm$ 0.6} \\
        crUCB (mod.)          & \colorABB{33.7 $\pm$ 1.8}{33.6 $\pm$ 1.7}{33.9 $\pm$ 2.2}{31.1 $\pm$ 2.3}{31.8 $\pm$ 2.2}{29.9 $\pm$ 3.0}{33.4 $\pm$ 2.7}{33.5 $\pm$ 1.5} & \colorABBclean{15.1 $\pm$ 1.0}{9.5 $\pm$ 0.3} \\
        \midrule
        \multicolumn{11}{c}{$\varepsilon = 0.4$} \\
        \midrule
        AT-DPT                & \colorABC{23.7 $\pm$ 1.8}{24.1 $\pm$ 2.0}{29.5 $\pm$ 3.2}{26.9 $\pm$ 1.8}{24.3 $\pm$ 1.1}{23.6 $\pm$ 2.0}{23.5 $\pm$ 0.8}{23.4 $\pm$ 1.7} & 		\colorABCclean{37.7 $\pm$ 3.1}{13.2 $\pm$ 0.7} \\
        AT-DPT (sub.\ 10\%) & \colorABC{30.6 $\pm$ 2.3}{30.9 $\pm$ 2.8}{35.9 $\pm$ 3.7}{32.8 $\pm$ 2.2}{30.9 $\pm$ 2.4}{29.7 $\pm$ 1.9}{29.9 $\pm$ 1.8}{30.3 $\pm$ 2.5} & 		\colorABCclean{41.6 $\pm$ 2.4}{17.4 $\pm$ 1.8} \\
        AT-DPT (sub.\ 20\%) & \colorABC{35.4 $\pm$ 2.4}{35.8 $\pm$ 3.2}{40.0 $\pm$ 3.9}{37.9 $\pm$ 2.4}{36.0 $\pm$ 2.7}{35.1 $\pm$ 3.3}{34.5 $\pm$ 2.8}{34.7 $\pm$ 2.9} & 		\colorABCclean{42.9 $\pm$ 2.9}{21.0 $\pm$ 2.9} \\
        AT-DPT (sub.\ 30\%) & \colorABC{41.2 $\pm$ 2.9}{41.9 $\pm$ 3.8}{45.7 $\pm$ 3.3}{43.6 $\pm$ 2.5}{41.6 $\pm$ 3.0}{41.3 $\pm$ 3.4}{42.0 $\pm$ 2.1}{41.0 $\pm$ 3.4} & 		\colorABCclean{47.8 $\pm$ 3.5}{25.9 $\pm$ 3.3} \\
        AT-DPT (rand.)        & \colorABC{39.3 $\pm$ 4.1}{40.3 $\pm$ 4.3}{43.4 $\pm$ 3.7}{42.1 $\pm$ 3.5}{40.0 $\pm$ 2.7}{42.3 $\pm$ 3.4}{40.6 $\pm$ 3.9}{40.3 $\pm$ 4.1} & 		\colorABCclean{38.3 $\pm$ 3.5}{20.6 $\pm$ 2.6} \\
        DPT                   & \colorABC{65.1 $\pm$ 8.0}{59.7 $\pm$ 5.1}{62.0 $\pm$ 8.5}{60.2 $\pm$ 7.2}{55.3 $\pm$ 7.8}{42.1 $\pm$ 6.4}{58.2 $\pm$ 7.1}{58.6 $\pm$ 7.7} & 		\colorABCclean{35.7 $\pm$ 3.7}{11.5 $\pm$ 0.5} \\
        AD [crUCB (m.)]       & \colorABC{55.2 $\pm$ 2.5}{56.1 $\pm$ 1.5}{57.0 $\pm$ 2.2}{53.9 $\pm$ 1.3}{55.3 $\pm$ 2.1}{50.5 $\pm$ 2.1}{54.5 $\pm$ 2.4}{52.7 $\pm$ 2.0} & 		\colorABCclean{33.1 $\pm$ 1.6}{16.1 $\pm$ 0.5} \\
        AD [crUCB (m.) c.]    & \colorABC{83.9 $\pm$ 4.6}{80.1 $\pm$ 4.4}{77.6 $\pm$ 3.8}{73.9 $\pm$ 3.2}{76.1 $\pm$ 3.4}{61.6 $\pm$ 2.9}{76.2 $\pm$ 3.7}{76.0 $\pm$ 3.2} & 		\colorABCclean{32.9 $\pm$ 1.8}{15.6 $\pm$ 0.5} \\
        AD [TS]               & \colorABC{94.8 $\pm$ 2.6}{84.5 $\pm$ 4.7}{82.5 $\pm$ 4.8}{75.0 $\pm$ 5.3}{78.7 $\pm$ 4.4}{46.3 $\pm$ 2.4}{80.7 $\pm$ 4.3}{81.6 $\pm$ 1.8} & 		\colorABCclean{35.5 $\pm$ 2.4}{9.1 $\pm$ 0.6} \\
        TS                    & \colorABC{107.1 $\pm$ 2.4}{97.8 $\pm$ 4.1}{94.1 $\pm$ 3.8}{94.9 $\pm$ 5.6}{90.8 $\pm$ 2.4}{49.2 $\pm$ 2.3}{92.1 $\pm$ 6.5}{92.4 $\pm$ 4.5} & 		\colorABCclean{34.4 $\pm$ 3.1}{8.7 $\pm$ 0.6} \\
        RTS ($\Cbar$ tuned)   & \colorABC{105.8 $\pm$ 2.4}{97.3 $\pm$ 4.3}{90.4 $\pm$ 4.4}{93.0 $\pm$ 4.6}{89.9 $\pm$ 3.9}{47.3 $\pm$ 3.1}{90.1 $\pm$ 5.1}{87.9 $\pm$ 2.7} & 		\colorABCclean{33.7 $\pm$ 2.0}{10.2 $\pm$ 0.4} \\
        RTS ($\Cbar$ unk.)    & \colorABC{113.9 $\pm$ 2.1}{107.9 $\pm$ 2.6}{104.3 $\pm$ 3.0}{104.7 $\pm$ 3.0}{103.3 $\pm$ 2.5}{74.3 $\pm$ 2.3}{103.9 $\pm$ 2.7}{104.7 $\pm$ 1.9} & 	\colorABCclean{62.3 $\pm$ 1.4}{48.9 $\pm$ 0.4} \\
        RTS ($\Cbar$ known)   & \colorABC{156.4 $\pm$ 1.9}{155.3 $\pm$ 1.9}{154.6 $\pm$ 1.7}{154.6 $\pm$ 2.0}{154.7 $\pm$ 2.0}{149.7 $\pm$ 2.0}{154.9 $\pm$ 1.9}{155.3 $\pm$ 2.0} & \colorABCclean{139.7 $\pm$ 1.9}{140.2 $\pm$ 1.6} \\
        UCB1.0                & \colorABC{105.6 $\pm$ 3.0}{95.3 $\pm$ 4.8}{91.1 $\pm$ 4.2}{90.3 $\pm$ 4.9}{88.7 $\pm$ 3.2}{46.6 $\pm$ 2.2}{91.8 $\pm$ 4.0}{91.4 $\pm$ 3.4} & 		\colorABCclean{39.2 $\pm$ 2.1}{16.0 $\pm$ 0.5} \\
        crUCB (orig.)         & \colorABC{148.6 $\pm$ 1.9}{147.7 $\pm$ 1.9}{147.9 $\pm$ 1.9}{147.2 $\pm$ 2.1}{147.8 $\pm$ 1.7}{145.1 $\pm$ 1.9}{147.5 $\pm$ 1.8}{148.1 $\pm$ 1.8} & \colorABCclean{141.2 $\pm$ 2.0}{141.4 $\pm$ 1.9} \\
        crUCB (low $\sigma_0$)& \colorABC{86.4 $\pm$ 2.7}{82.9 $\pm$ 3.8}{82.5 $\pm$ 3.1}{83.9 $\pm$ 3.9}{80.9 $\pm$ 4.1}{64.7 $\pm$ 3.8}{82.5 $\pm$ 4.0}{84.4 $\pm$ 4.5} & 		\colorABCclean{31.1 $\pm$ 1.5}{15.0 $\pm$ 0.4} \\
        crUCB (mod.)          & \colorABC{85.7 $\pm$ 3.7}{84.3 $\pm$ 2.2}{81.9 $\pm$ 4.4}{82.1 $\pm$ 3.7}{79.3 $\pm$ 2.9}{63.9 $\pm$ 2.2}{80.1 $\pm$ 3.7}{81.9 $\pm$ 4.9} & 		\colorABCclean{31.8 $\pm$ 1.4}{15.8 $\pm$ 0.5} \\
        \bottomrule
        \vspace*{-1.5em}
    \end{tabular}
    \end{adjustbox}
    \end{footnotesize}
\end{table*}
\clearpage
\renewcommand{\floatpagefraction}{1}
\subsection{Different Budgets}
\vspace*{-0.5em}

\begin{figure}[!htb]
    \centering
    \includegraphics[width=0.95\linewidth]{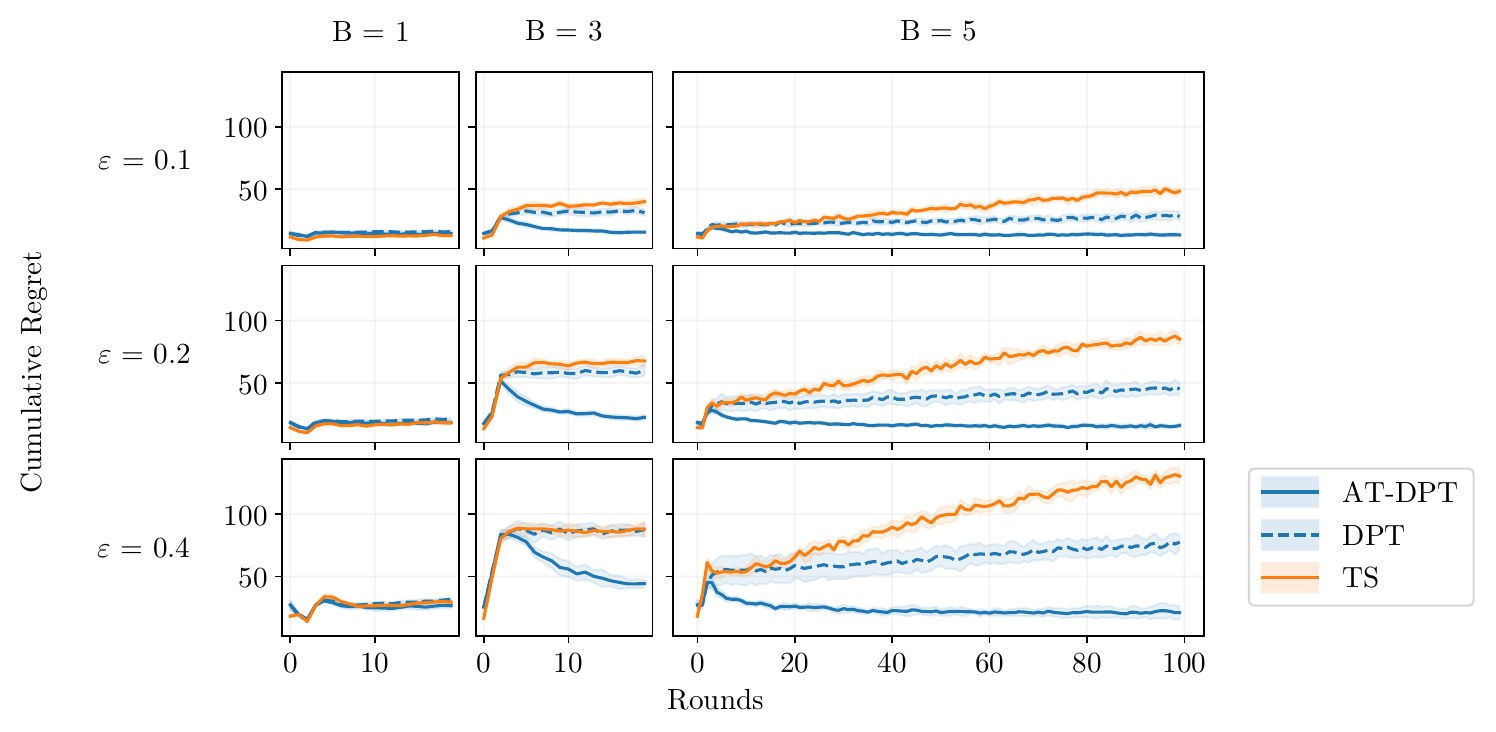}
    \caption{
        A study of the effect of the budget $B$ on the regret in the bandit setting.
        We run the experiments for $B=5$ for more rounds to observe convergence.
        We observe that a larger budget for the attacker leads to a higher regret for TS and DPT, although adversarial training for AT-DPT helps it learn to recover from the attack.
    }
    \label{fig:budgets}
\end{figure}

\setlength{\tabcolsep}{-2pt}
\begin{table*}[!htb]
    \footnotesize
    \centering
    \caption{Comparison of the cumulative regret (lower is better) of the different algorithms under different attackers trained for 20 rounds, in the bandit setting, \textbf{across different budgets}. Mean and 95\% confidence interval (2$\times$SEM) over 10 experiment replications. Fraction of steps poisoned $\varepsilon = 0.4$.}
    \begin{tabular}{lcccccc|cc}
        \toprule
        \multirow{2}{*}{Algorithm} & \multicolumn{6}{c|}{Attacker Target} & \multirow[b]{2}{*}{\shortstack[c]{Unif.\ Rand.\\Attack}} & \multirow{2}{*}{Clean Env.} \\
        & AT-DPT & DPT & TS & RTS ($\Cbar$ t.) & UCB1.0 & crUCB (m.) \\
        \midrule
        \multicolumn{9}{c}{$B_\text{train}=1,\; B_\text{test} = 1$} \\
        \midrule
        AT-DPT     & \colorABbudget{29.3 $\pm$ 2.0}{28.2 $\pm$ 2.2}{28.4 $\pm$ 1.5}{28.6 $\pm$ 1.8}{28.5 $\pm$ 2.1}{28.5 $\pm$ 2.5}{31.7 $\pm$ 2.4}{14.7 $\pm$ 1.9} \\
        DPT        & \colorABbudget{35.0 $\pm$ 2.2}{31.6 $\pm$ 1.7}{33.3 $\pm$ 2.1}{32.8 $\pm$ 1.6}{32.7 $\pm$ 2.0}{33.0 $\pm$ 3.1}{28.2 $\pm$ 1.9}{11.5 $\pm$ 0.5} \\
        TS         & \colorABbudget{34.3 $\pm$ 2.0}{29.9 $\pm$ 1.2}{30.1 $\pm$ 1.0}{31.8 $\pm$ 1.3}{30.3 $\pm$ 1.7}{30.8 $\pm$ 2.3}{26.7 $\pm$ 2.2}{8.7 $\pm$ 0.6} \\
        RTS ($\Cbar$ tuned)& \colorABbudget{34.3 $\pm$ 1.9}{30.9 $\pm$ 1.4}{30.6 $\pm$ 1.8}{31.2 $\pm$ 1.6}{30.2 $\pm$ 0.9}{30.7 $\pm$ 1.6}{27.1 $\pm$ 1.6}{10.2 $\pm$ 0.4} \\
        UCB1.0     & \colorABbudget{40.1 $\pm$ 1.6}{36.2 $\pm$ 1.4}{37.2 $\pm$ 1.2}{37.2 $\pm$ 1.8}{36.6 $\pm$ 1.7}{36.2 $\pm$ 1.5}{31.4 $\pm$ 1.5}{16.0 $\pm$ 0.5} \\
        crUCB (mod.)&\colorABbudget{37.1 $\pm$ 1.4}{33.6 $\pm$ 1.5}{34.3 $\pm$ 1.5}{33.1 $\pm$ 1.6}{33.3 $\pm$ 1.8}{34.6 $\pm$ 1.9}{26.9 $\pm$ 0.9}{15.8 $\pm$ 0.5} \\
        \midrule
        \multicolumn{9}{c}{$B_\text{train}=1,\; B_\text{test} = 3$} \\
        \midrule
        AT-DPT     & \colorABbudget{28.5 $\pm$ 1.6}{29.3 $\pm$ 1.4}{34.1 $\pm$ 2.9}{31.2 $\pm$ 2.4}{28.9 $\pm$ 2.1 }{27.8 $\pm$ 1.7}{38.5 $\pm$ 6.4}{14.7 $\pm$ 1.9} \\
        DPT        & \colorABbudget{63.5 $\pm$ 6.6}{58.2 $\pm$ 5.4}{61.4 $\pm$ 7.5}{59.8 $\pm$ 7.7}{57.4 $\pm$ 8.1 }{58.0 $\pm$ 6.5}{34.1 $\pm$ 5.5}{11.5 $\pm$ 0.5} \\
        TS         & \colorABbudget{107.1 $\pm$ 2.4}{97.8 $\pm$ 4.1}{94.1 $\pm$ 3.8}{94.9 $\pm$ 5.6}{90.8 $\pm$ 2.4}{92.4 $\pm$ 4.5}{31.6 $\pm$ 5.6}{8.7 $\pm$ 0.6} \\
        RTS ($\Cbar$ tuned)& \colorABbudget{105.8 $\pm$ 2.4}{97.3 $\pm$ 4.3}{90.4 $\pm$ 4.4}{93.0 $\pm$ 4.6}{89.9 $\pm$ 3.9}{87.9 $\pm$ 2.7}{31.4 $\pm$ 5.6}{10.2 $\pm$ 0.4} \\
        UCB1.0     & \colorABbudget{105.6 $\pm$ 3.0}{95.3 $\pm$ 4.8}{91.1 $\pm$ 4.2}{90.3 $\pm$ 4.9}{88.7 $\pm$ 3.2}{91.4 $\pm$ 3.4}{34.3 $\pm$ 5.7}{16.0 $\pm$ 0.5} \\
        crUCB (mod.)&\colorABbudget{85.7 $\pm$ 3.7}{84.3 $\pm$ 2.2}{81.9 $\pm$ 4.4}{82.1 $\pm$ 3.7}{79.3 $\pm$ 2.9 }{81.9 $\pm$ 4.9}{28.9 $\pm$ 3.6}{15.8 $\pm$ 0.5} \\
        \midrule
        \multicolumn{9}{c}{$B_\text{train}=1,\; B_\text{test} = 5$} \\
        \midrule
        AT-DPT     & \colorABbudget{34.9 $\pm$ 1.7}{35.7 $\pm$ 2.3}{40.5 $\pm$ 3.5}{37.7 $\pm$ 2.6}{35.5 $\pm$ 2.1  }{35.0 $\pm$ 2.9}{69.8 $\pm$ 12.6}{14.7 $\pm$ 1.9} \\
        DPT        & \colorABbudget{58.7 $\pm$ 8.4}{56.1 $\pm$ 11.2}{61.7 $\pm$ 9.3}{59.2 $\pm$ 10.0}{57.1 $\pm$ 9.7}{56.6 $\pm$ 10.6}{58.3 $\pm$ 10.0}{11.5 $\pm$ 0.5} \\
        TS         & \colorABbudget{65.0 $\pm$ 5.5}{59.5 $\pm$ 6.1}{65.3 $\pm$ 4.1}{60.9 $\pm$ 6.1}{53.7 $\pm$ 3.8  }{59.8 $\pm$ 4.3}{54.9 $\pm$ 10.4}{8.7 $\pm$ 0.6} \\
        RTS ($\Cbar$ tuned)& \colorABbudget{61.5 $\pm$ 5.3}{57.9 $\pm$ 4.9}{62.3 $\pm$ 5.3}{61.2 $\pm$ 6.1}{53.1 $\pm$ 4.7  }{56.8 $\pm$ 3.7}{53.4 $\pm$ 10.6}{10.2 $\pm$ 0.4} \\
        UCB1.0     & \colorABbudget{60.1 $\pm$ 4.6}{53.2 $\pm$ 4.5}{64.4 $\pm$ 4.4}{58.5 $\pm$ 5.5}{50.5 $\pm$ 3.9  }{52.2 $\pm$ 3.9}{59.5 $\pm$ 12.1}{16.0 $\pm$ 0.5} \\
        crUCB (mod.)&\colorABbudget{79.9 $\pm$ 2.9}{80.2 $\pm$ 3.6}{82.0 $\pm$ 4.4}{81.6 $\pm$ 3.2}{75.3 $\pm$ 5.1  }{80.0 $\pm$ 3.4}{41.9 $\pm$ 6.5}{15.8 $\pm$ 0.5} \\
        \bottomrule
    \end{tabular}
    \label{tab:bandit_compare_budget}
\end{table*}
\clearpage

We present experiments showing how AT-DPT trains on different attack budgets $B$ in \Cref{fig:budgets}.
In all three variants tested ($B=1, 3, 5$) AT-DPT seems to recover within approx.\ 20 rounds and performs well.

While we test on out-of-distribution attacks in all experiments, meaning the test-phase attacks differ from the train-phase attacks, we perform an orthogonal out-of-distribution study on the impact of the budget during training ($B_\text{train}$) and evaluation ($B_\text{test}$) in \Cref{tab:bandit_compare_budget}.
This shows how training on a weaker attacker ($B=1$) transfers to stronger attacks ($B=3$, $B=5$).
The setup is otherwise the same as in the other tables.
The results seem to show that there is evidence that AT-DPT transfers robustness to stronger attacks.
That is, AT-DPT trained with a weaker type of attack learns to be robust against a stronger attacker.
We think it would be interesting to consider generalization to different attack vectors, such as observation or action poisoning, in future work, and there is potential to apply an approach similar to ours.

\subsection{Bandit setting, adaptive attack}

\Cref{tab:bandit_adaptive_full} presents the full set of results comparing adaptive and non-adaptive attacks.
% In these results we would like to highlight the difference between the non-adaptive attacker performance, and adaptive attacker performance.
The adaptive attacker columns in the table highlight, that these attacks work much better from the attacker's perspective, i.e., the attacks increase regret by a larger margin than in the non-adaptive case.
We note that in both cases the regret of AT-DPT is low, meaning it is working well.

\setlength{\tabcolsep}{0pt}
\begin{table*}[!htb]
    \footnotesize
    \centering
    \caption{Comparison of the cumulative regret (lower is better) of \textbf{adaptive and non-adaptive attackers} in the bandit setting. Attackers trained for 400 rounds. Mean and 95\% confidence interval (2$\times$SEM) over 10 experiment replications. Attack budget $B = 3$. $^*$ We use tuned versions of RTS and crUCB which outperform the base versions -- see \Cref{app:baselines} for details.
    }
    % \hspace*{-4.1em}
    \begin{adjustbox}{bgcolor=white}
    \begin{tabular}{lcc|cc|cc}
        \toprule
        \multirow{3}{*}{Algorithm} & \multicolumn{4}{c|}{Attacker Target} & \multirow[b]{3}{*}{\shortstack[c]{Unif.\ Rand.\\Attack}} & \multirow{3}{*}{Clean Env.} \\
        & \multicolumn{2}{c|}{Adaptive} & \multicolumn{2}{c|}{Non-adaptive} \\
        & AT-DPT & TS & AT-DPT & TS & \\
        \midrule
        \multicolumn{7}{c}{$\varepsilon = 0.1$} \\
        \midrule
        AT-DPT (against adaptive)     & \colorBAdaptiveVsNonA{21.5 $\pm$ 4.2}{22.6 $\pm$ 5.5}{19.6 $\pm$ 3.8}{19.6 $\pm$ 4.2}{21.4 $\pm$ 9.4}{16.7 $\pm$ 5.1} \\
        AT-DPT (against non-adaptive) & \colorBAdaptiveVsNonA{44.6 $\pm$ 17.7}{39.7 $\pm$ 15.4}{14.1 $\pm$ 0.7}{14.2 $\pm$ 0.6}{14.7 $\pm$ 1.0}{11.8 $\pm$ 1.1} \\
        DPT                           & \colorBAdaptiveVsNonA{60.2 $\pm$ 18.5}{47.1 $\pm$ 10.1}{22.2 $\pm$ 1.1}{21.9 $\pm$ 1.5}{15.1 $\pm$ 1.0}{12.1 $\pm$ 0.8} \\
        TS                            & \colorBAdaptiveVsNonA{102.8 $\pm$ 27.1}{91.0 $\pm$ 29.3}{27.8 $\pm$ 1.7}{27.4 $\pm$ 1.6}{11.9 $\pm$ 0.7}{9.1 $\pm$ 0.7} \\
        RTS ($\Cbar$ tuned)           & \colorBAdaptiveVsNonA{101.5 $\pm$ 26.2}{91.7 $\pm$ 29.0}{26.1 $\pm$ 2.1}{26.4 $\pm$ 2.1}{12.9 $\pm$ 0.7}{10.5 $\pm$ 0.6} \\
        UCB                           & \colorBAdaptiveVsNonA{103.9 $\pm$ 26.0}{94.0 $\pm$ 27.2}{28.8 $\pm$ 1.1}{29.2 $\pm$ 1.1}{18.1 $\pm$ 0.6}{16.0 $\pm$ 0.4} \\
        crUCB (mod.)                  & \colorBAdaptiveVsNonA{67.2 $\pm$ 17.1}{53.3 $\pm$ 15.8}{18.9 $\pm$ 1.3}{19.6 $\pm$ 1.5}{10.8 $\pm$ 0.5}{9.2 $\pm$ 0.3} \\
        \midrule
        \multicolumn{7}{c}{$\varepsilon = 0.2$} \\
        \midrule
        AT-DPT (against adaptive)     & \colorBAdaptiveVsNonB{26.6 $\pm$ 5.3}{29.1 $\pm$ 8.5}{25.9 $\pm$ 5.1}{27.0 $\pm$ 4.5}{27.2 $\pm$ 10.9}{18.8 $\pm$ 7.6} \\
        AT-DPT (against non-adaptive) & \colorBAdaptiveVsNonB{54.7 $\pm$ 15.5}{51.8 $\pm$ 11.8}{17.5 $\pm$ 0.9}{19.4 $\pm$ 1.3}{20.5 $\pm$ 1.9}{12.4 $\pm$ 1.3} \\
        DPT                           & \colorBAdaptiveVsNonB{71.3 $\pm$ 20.4}{61.6 $\pm$ 17.0}{34.6 $\pm$ 3.6}{35.1 $\pm$ 3.1}{22.0 $\pm$ 1.9}{12.1 $\pm$ 0.8} \\
        TS                            & \colorBAdaptiveVsNonB{74.9 $\pm$ 24.6}{91.6 $\pm$ 35.0}{51.6 $\pm$ 2.6}{51.8 $\pm$ 3.6}{18.8 $\pm$ 1.5}{9.1 $\pm$ 0.7} \\
        RTS ($\Cbar$ tuned)           & \colorBAdaptiveVsNonB{75.5 $\pm$ 24.6}{92.2 $\pm$ 34.2}{49.5 $\pm$ 2.9}{49.3 $\pm$ 2.7}{19.3 $\pm$ 1.2}{10.5 $\pm$ 0.6} \\
        UCB                           & \colorBAdaptiveVsNonB{76.8 $\pm$ 22.4}{92.4 $\pm$ 30.9}{51.6 $\pm$ 2.7}{49.4 $\pm$ 1.9}{23.6 $\pm$ 1.1}{16.0 $\pm$ 0.4} \\
        crUCB (mod.)                  & \colorBAdaptiveVsNonB{55.1 $\pm$ 16.5}{61.8 $\pm$ 25.0}{34.6 $\pm$ 1.8}{34.4 $\pm$ 1.2}{14.1 $\pm$ 0.8}{9.5 $\pm$ 0.5} \\
        \midrule
        \multicolumn{7}{c}{$\varepsilon = 0.4$} \\
        \midrule
        AT-DPT (against adaptive)     & \colorBAdaptiveVsNonC{37.1 $\pm$ 6.6}{36.4 $\pm$ 9.4}{38.0 $\pm$ 6.4}{42.6 $\pm$ 6.7}{41.4 $\pm$ 7.3}{21.3 $\pm$ 9.0} \\
        AT-DPT (against non-adaptive) & \colorBAdaptiveVsNonC{88.1 $\pm$ 20.0}{81.0 $\pm$ 11.2}{22.8 $\pm$ 1.6}{29.8 $\pm$ 2.2}{39.7 $\pm$ 3.8}{13.8 $\pm$ 1.2} \\
        DPT                           & \colorBAdaptiveVsNonC{97.9 $\pm$ 18.6}{82.1 $\pm$ 20.7}{61.6 $\pm$ 8.0}{61.6 $\pm$ 6.6}{37.3 $\pm$ 3.5}{12.1 $\pm$ 0.8} \\
        TS                            & \colorBAdaptiveVsNonC{90.2 $\pm$ 21.9}{104.2 $\pm$ 26.7}{106.3 $\pm$ 5.5}{94.3 $\pm$ 4.8}{34.1 $\pm$ 2.5}{9.1 $\pm$ 0.7} \\
        RTS ($\Cbar$ tuned)           & \colorBAdaptiveVsNonC{90.5 $\pm$ 21.3}{103.6 $\pm$ 26.8}{104.5 $\pm$ 5.5}{90.9 $\pm$ 4.2}{34.5 $\pm$ 2.4}{10.5 $\pm$ 0.6} \\
        UCB                           & \colorBAdaptiveVsNonC{94.3 $\pm$ 22.4}{103.9 $\pm$ 28.4}{101.3 $\pm$ 5.0}{87.8 $\pm$ 4.4}{38.2 $\pm$ 1.6}{16.0 $\pm$ 0.4} \\
        crUCB (mod.)                  & \colorBAdaptiveVsNonC{85.1 $\pm$ 23.5}{79.6 $\pm$ 29.4}{88.4 $\pm$ 4.4}{79.9 $\pm$ 4.7}{32.0 $\pm$ 1.7}{15.8 $\pm$ 0.3} \\
        \bottomrule
    \end{tabular}
    \end{adjustbox}
    \label{tab:bandit_adaptive_full}
\end{table*}

\clearpage
\subsection{Linear bandit setting}

\Cref{tab:linear_bandit_results_full} presents the full results from the linear bandit setting.
As described in \Cref{app:crlinucb}, CRLinUCBv1 and CRLinUCBv3 performed worse than the tuned version CRLinUCBv2.
This is indicated by their poor performance on the clean and uniform random attack cases.

\setlength{\tabcolsep}{-2pt}
\begin{table*}[!htb]
    % \footnotesize
    \centering
    \begin{footnotesize}
    \caption{Comparison of the cumulative regret (lower is better) of the different algorithms under different attackers in the \textbf{linear bandit setting}. Mean and 95\% confidence interval (2$\times$SEM) over 10 experiment replications. Attack budget $B = 3$.}
    \label{tab:linear_bandit_results_full}
    % \vspace*{-5pt}
    % \hspace*{-4.3em}
    \begin{adjustbox}{bgcolor=white}
    \begin{tabular}{lcccccc|cc}
        \toprule
        \multirow{2}{*}{Algorithm} & \multicolumn{6}{c|}{Attacker Target} & \multirow[b]{2}{*}{\shortstack[c]{Unif.\ Rand.\\Attack}} & \multirow{2}{*}{Clean Env.} \\
        & AT-DPT & DPT   & LinUCB & CRLinUCBv1 & CRLinUCBv2 & CRLinUCBv3 \\
        \midrule
        \multicolumn{9}{c}{$\varepsilon = 0.1$} \\
        \midrule
        AT-DPT         & \colorALBC{2.55 $\pm$ 0.88}{2.02 $\pm$ 0.92}{2.28 $\pm$ 0.90}{2.44 $\pm$ 0.96}{1.55 $\pm$ 0.98}{1.85 $\pm$ 0.94}{4.60 $\pm$ 0.98}{3.89 $\pm$ 0.86} \\
        DPT            & \colorALBC{14.50 $\pm$ 2.30}{14.42 $\pm$ 2.48}{14.62 $\pm$ 2.34}{14.06 $\pm$ 2.72}{14.02 $\pm$ 2.74}{13.19 $\pm$ 2.36}{5.23 $\pm$ 1.02}{3.35 $\pm$ 0.84} \\
        LinUCB         & \colorALBC{10.57 $\pm$ 2.00}{7.92 $\pm$ 1.24}{7.56 $\pm$ 1.20}{9.65 $\pm$ 1.78}{8.04 $\pm$ 1.52}{9.23 $\pm$ 1.72}{4.18 $\pm$ 0.90}{3.51 $\pm$ 0.88} \\
        CRLinUCBv1    & \colorALBC{104.00 $\pm$ 7.06}{103.24 $\pm$ 7.00}{103.11 $\pm$ 7.04}{103.03 $\pm$ 6.98}{102.56 $\pm$ 7.02}{102.97 $\pm$ 7.00}{102.95 $\pm$ 7.00}{110.85 $\pm$ 7.78} \\
        CRLinUCBv2    & \colorALBC{7.85 $\pm$ 1.58}{7.99 $\pm$ 1.60}{10.16 $\pm$ 2.00}{10.94 $\pm$ 2.34}{7.82 $\pm$ 1.92}{9.07 $\pm$ 2.48}{3.16 $\pm$ 0.96}{2.94 $\pm$ 0.78} \\
        CRLinUCBv3    & \colorALBC{17.66 $\pm$ 1.18}{17.34 $\pm$ 1.18}{17.40 $\pm$ 1.24}{17.86 $\pm$ 1.20}{16.99 $\pm$ 1.14}{16.58 $\pm$ 1.12}{13.55 $\pm$ 0.96}{34.08 $\pm$ 1.66} \\
        \midrule
        \multicolumn{9}{c}{$\varepsilon = 0.2$} \\
        \midrule
        AT-DPT         & \colorALBC{1.37 $\pm$ 0.94}{1.20 $\pm$ 0.92}{1.67 $\pm$ 0.96}{1.30 $\pm$ 0.96}{2.42 $\pm$ 0.94}{2.00 $\pm$ 0.92}{4.80 $\pm$ 0.98}{3.89 $\pm$ 0.86} \\
        DPT            & \colorALBC{33.65 $\pm$ 4.14}{35.49 $\pm$ 4.66}{32.23 $\pm$ 3.96}{35.29 $\pm$ 4.24}{33.67 $\pm$ 4.02}{33.15 $\pm$ 3.84}{5.91 $\pm$ 1.12}{3.35 $\pm$ 0.84} \\
        LinUCB         & \colorALBC{19.11 $\pm$ 2.56}{15.79 $\pm$ 2.32}{18.80 $\pm$ 2.62}{21.31 $\pm$ 3.16}{16.95 $\pm$ 2.68}{19.52 $\pm$ 2.62}{4.37 $\pm$ 0.98}{3.51 $\pm$ 0.88} \\
        CRLinUCBv1    & \colorALBC{100.19 $\pm$ 6.88}{99.73 $\pm$ 6.74}{99.35 $\pm$ 6.82}{99.41 $\pm$ 6.88}{100.48 $\pm$ 6.90}{100.34 $\pm$ 6.88}{107.34 $\pm$ 7.44}{110.85 $\pm$ 7.78} \\
        CRLinUCBv2    & \colorALBC{16.42 $\pm$ 2.76}{13.97 $\pm$ 2.46}{16.56 $\pm$ 2.46}{22.27 $\pm$ 3.66}{16.02 $\pm$ 2.64}{18.45 $\pm$ 2.78}{3.41 $\pm$ 1.02}{2.94 $\pm$ 0.78} \\
        CRLinUCBv3    & \colorALBC{31.53 $\pm$ 1.76}{28.93 $\pm$ 1.62}{28.66 $\pm$ 1.58}{30.44 $\pm$ 1.80}{30.02 $\pm$ 1.64}{30.20 $\pm$ 1.74}{19.42 $\pm$ 1.08}{34.08 $\pm$ 1.66} \\
        \midrule
        \multicolumn{9}{c}{$\varepsilon = 0.4$} \\
        \midrule
        AT-DPT         & \colorALBC{2.49 $\pm$ 1.06}{2.50 $\pm$ 1.08}{2.83 $\pm$ 1.10}{2.93 $\pm$ 1.06}{1.79 $\pm$ 1.02}{2.16 $\pm$ 1.10}{5.33 $\pm$ 1.16}{3.89 $\pm$ 0.86} \\
        DPT            & \colorALBC{70.29 $\pm$ 7.32}{71.42 $\pm$ 7.46}{70.83 $\pm$ 7.76}{69.49 $\pm$ 6.88}{63.84 $\pm$ 7.18}{73.45 $\pm$ 6.82}{6.62 $\pm$ 1.30}{3.35 $\pm$ 0.84} \\
        LinUCB         & \colorALBC{37.69 $\pm$ 4.46}{35.93 $\pm$ 3.86}{35.22 $\pm$ 4.14}{49.39 $\pm$ 5.12}{34.82 $\pm$ 4.36}{39.97 $\pm$ 4.50}{5.21 $\pm$ 1.16}{3.51 $\pm$ 0.88} \\
        CRLinUCBv1    & \colorALBC{108.12 $\pm$ 6.96}{107.54 $\pm$ 7.00}{108.04 $\pm$ 6.98}{107.84 $\pm$ 6.98}{106.79 $\pm$ 6.98}{107.13 $\pm$ 6.90}{109.46 $\pm$ 7.64}{110.85 $\pm$ 7.78} \\
        CRLinUCBv2    & \colorALBC{37.45 $\pm$ 4.76}{33.03 $\pm$ 4.00}{35.56 $\pm$ 4.26}{46.23 $\pm$ 5.46}{35.36 $\pm$ 4.80}{37.75 $\pm$ 4.80}{5.12 $\pm$ 1.48}{2.94 $\pm$ 0.78} \\
        CRLinUCBv3    & \colorALBC{53.23 $\pm$ 3.02}{51.76 $\pm$ 2.84}{53.31 $\pm$ 2.98}{54.45 $\pm$ 3.08}{49.13 $\pm$ 2.66}{51.43 $\pm$ 2.78}{28.34 $\pm$ 1.56}{34.08 $\pm$ 1.66} \\
        \bottomrule
    \end{tabular}
    \end{adjustbox}
    \end{footnotesize}
\end{table*}

\clearpage
\subsection{MDP setting}\label{app:mdp_setting}

% In \Cref{tab:mdp_megatable_epse0.8_full} we present the full set of results for the Darkroom2 environment.

\begin{table*}[!h]
    \vspace*{-0.5em}
    \footnotesize
    \centering
    \caption{Comparison of the average episode reward (higher is better) of the different algorithms under different attackers trained for 300 rounds (5 rounds for Q-learning and NPG) in the \textbf{Darkroom2 environment} (5$\times$5 grid). Mean and 95\% confidence interval (2$\times$SEM) over 10 experiment replications. Attack budget $B = 10$. $^\S$ NPG and Q-learning require multiple episodes of online learning to converge to a stable policy; we run them for 100 episodes before evaluating their performance.}
    \vspace*{-0.5em}
    \begin{tabular}{lcccc|cc}
        \toprule
        \multirow{2}{*}{Algorithm} & \multicolumn{4}{c|}{Attacker Target} & \multirow[b]{2}{*}{\shortstack[c]{Unif.\ Rand.\\Attack}} & \multirow{2}{*}{Clean Env.} \\
        & AT-DPT & DPT & NPG & Q-learning \\
        \midrule
        \multicolumn{7}{c}{$\varepsilon = 0.1$} \\
        \midrule
        AT-DPT          & \colorMC{269.9 $\pm$ 16.3}{266.0 $\pm$ 20.2}{262.3 $\pm$ 16.6}{258.9 $\pm$ 20.2}{271.4 $\pm$ 20.1}{272.7 $\pm$ 18.3} \\
        AT-DPT (sub.\ 10\%)	& \colorMC{262.0 $\pm$ 23.1}{267.6 $\pm$ 22.9}{261.3 $\pm$ 24.3}{263.3 $\pm$ 25.5}{271.6 $\pm$ 22.7}{274.8 $\pm$ 21.4} \\
        AT-DPT (sub.\ 20\%)	& \colorMC{248.4 $\pm$ 21.0}{246.6 $\pm$ 19.6}{244.4 $\pm$ 19.8}{242.5 $\pm$ 22.3}{249.2 $\pm$ 23.0}{247.1 $\pm$ 21.7} \\
        AT-DPT (sub.\ 30\%)	& \colorMC{217.1 $\pm$ 18.0}{218.8 $\pm$ 18.4}{213.3 $\pm$ 18.3}{215.6 $\pm$ 17.1}{222.1 $\pm$ 18.9}{221.6 $\pm$ 16.5} \\
        DPT             & \colorMC{236.8 $\pm$ 9.7}{199.0 $\pm$ 10.4}{224.8 $\pm$ 12.4}{222.6 $\pm$ 6.8}{277.4 $\pm$ 7.1}{306.8 $\pm$ 7.1} \\
        NPG$^\S$        & \colorMC{241.9 $\pm$ 6.6}{248.1 $\pm$ 6.3}{247.7 $\pm$ 7.1}{243.3 $\pm$ 5.5}{246.0 $\pm$ 6.9}{241.7 $\pm$ 7.5} \\
        Q-learning$^\S$ & \colorMC{280.1 $\pm$ 5.5}{281.1 $\pm$ 5.2}{248.5 $\pm$ 38.6}{264.1 $\pm$ 18.1}{266.5 $\pm$ 15.4}{266.0 $\pm$ 14.8} \\
        \midrule
        \multicolumn{7}{c}{$\varepsilon = 0.2$} \\
        \midrule
        AT-DPT          & \colorMC{261.0 $\pm$ 14.6}{271.7 $\pm$ 15.5}{257.7 $\pm$ 16.7}{258.0 $\pm$ 19.3}{270.1 $\pm$ 17.8}{279.9 $\pm$ 20.0} \\
        AT-DPT (sub.\ 10\%) & \colorMC{252.9 $\pm$ 15.4}{259.8 $\pm$ 12.3}{253.9 $\pm$ 16.1}{255.3 $\pm$ 14.5}{267.4 $\pm$ 15.0}{275.0 $\pm$ 15.6} \\
        AT-DPT (sub.\ 20\%) & \colorMC{230.7 $\pm$ 26.9}{233.7 $\pm$ 25.5}{229.4 $\pm$ 25.5}{231.7 $\pm$ 27.2}{238.1 $\pm$ 28.1}{241.9 $\pm$ 32.0} \\
        AT-DPT (sub.\ 30\%) & \colorMC{201.5 $\pm$ 24.5}{205.6 $\pm$ 24.2}{195.9 $\pm$ 21.0}{192.3 $\pm$ 27.0}{202.3 $\pm$ 23.5}{212.1 $\pm$ 23.9} \\
        DPT             & \colorMC{229.6 $\pm$ 7.1}{171.9 $\pm$ 11.7}{215.0 $\pm$ 8.1}{217.4 $\pm$ 9.5}{273.8 $\pm$ 7.8}{306.8 $\pm$ 7.1} \\
        NPG$^\S$        & \colorMC{244.1 $\pm$ 7.5}{244.4 $\pm$ 6.7}{239.5 $\pm$ 9.4}{241.2 $\pm$ 9.4}{248.9 $\pm$ 8.0}{241.7 $\pm$ 7.5} \\
        Q-learning$^\S$ & \colorMC{240.3 $\pm$ 5.4}{251.2 $\pm$ 7.3}{236.8 $\pm$ 5.8}{246.2 $\pm$ 6.0}{244.7 $\pm$ 6.8}{241.7 $\pm$ 8.2} \\
        \midrule
        \multicolumn{7}{c}{$\varepsilon = 0.4$} \\
        \midrule
        AT-DPT     & \colorMC{242.2 $\pm$ 11.9}{267.5 $\pm$ 10.5}{241.7 $\pm$ 10.2}{239.1 $\pm$ 8.8}{258.2 $\pm$ 11.8}{267.4 $\pm$ 15.1} \\
        AT-DPT (sub.\ 10\%) & \colorMC{225.2 $\pm$ 17.8}{245.5 $\pm$ 17.9}{226.1 $\pm$ 16.6}{222.7 $\pm$ 17.6}{241.3 $\pm$ 17.8}{250.2 $\pm$ 18.8} \\
        AT-DPT (sub.\ 20\%) & \colorMC{199.4 $\pm$ 20.4}{219.3 $\pm$ 23.6}{200.5 $\pm$ 22.2}{200.0 $\pm$ 22.3}{208.5 $\pm$ 25.8}{219.1 $\pm$ 26.3} \\
        AT-DPT (sub.\ 30\%) & \colorMC{178.6 $\pm$ 18.7}{195.5 $\pm$ 23.9}{178.8 $\pm$ 21.0}{176.2 $\pm$ 20.7}{182.6 $\pm$ 17.8}{190.5 $\pm$ 17.6} \\
        DPT        & \colorMC{216.1 $\pm$ 11.0}{143.5 $\pm$ 11.0}{202.6 $\pm$ 7.4}{205.9 $\pm$ 7.8}{266.2 $\pm$ 8.1}{306.8 $\pm$ 7.1} \\
        NPG$^\S$ & \colorMC{237.2 $\pm$ 6.7}{243.7 $\pm$ 7.9}{228.9 $\pm$ 4.0}{228.1 $\pm$ 8.1}{235.3 $\pm$ 8.2}{241.7 $\pm$ 7.5} \\
        Q-learning$^\S$ & \colorMC{198.1 $\pm$ 3.7}{238.6 $\pm$ 6.0}{215.4 $\pm$ 7.6}{224.7 $\pm$ 7.3}{229.0 $\pm$ 7.2}{225.6 $\pm$ 5.4} \\
        \bottomrule
    \end{tabular}
    \label{tab:mdp_megatable_epse0.8_full}
\end{table*}

\clearpage
For the PPO baseline in \Cref{tab:miniworld,tab:miniworld_full} we use the \texttt{cleanrl} implementation \citep{huang2022cleanrl}.
% In \Cref{tab:miniworld_full} we additionally show experiments from the Miniworld environment \citep{chevalier2023miniworld}, a 3D environment to evaluate visual navigation from images (25 $\times$ 25 pixels).
% We follow a similar setup as in the original DPT paper \citep{lee2023dpt}.
% The environment consists of four boxes of different colors, and one of those is chosen as the goal box, unknown to the agent.
% The agent receives a reward of $+1$ when stood next to the goal box.
% The episode is $H=250$ steps long.
% For the PPO baseline we use the \texttt{cleanrl} implementation \citep{huang2022cleanrl}.
{
\setlength{\textfloatsep}{5pt}
\begin{table*}[!h]
    \footnotesize
    \centering
    \caption{
        Comparison of the average episode reward (higher is better) of the different algorithms under different attackers trained for 100 rounds in the \textbf{Miniworld environment}.
        Mean and 95\% confidence interval (2$\times$SEM) over 10 runs.
        Attack budget $B = 5$.
        $^\S$ PPO requires multiple episodes of online learning to converge to a stable policy; we run it for 100 episodes before evaluating the performance.
    }
    \vspace*{-0.5em}
    \begin{tabular}{lcc|cc}
        \toprule
        \multirow{2}{*}{Algorithm} & \multicolumn{2}{c|}{Attacker Target} & \multirow[b]{2}{*}{\shortstack[c]{Unif.\ Rand.\\Attack}} & \multirow{2}{*}{Clean Env.} \\
        & AT-DPT & DPT \\
        \midrule
        \multicolumn{5}{c}{$\varepsilon = 0.1$} \\
        \midrule
        AT-DPT          & \colorMwA{111.1 $\pm$ 11.9}{114.1 $\pm$ 13.0}{110.1 $\pm$ 16.0}{123.9 $\pm$ 16.7} \\
        DPT             & \colorMwA{93.2 $\pm$ 12.4}{92.8 $\pm$ 14.2}{103.1 $\pm$ 12.8}{110.0 $\pm$ 14.7} \\
        PPO$^\S$        & \colorMwA{117.9 $\pm$ 8.4}{115.5 $\pm$ 4.1}{101.5 $\pm$ 6.2}{123.5 $\pm$ 8.1} \\
        \midrule
        \multicolumn{5}{c}{$\varepsilon = 0.2$} \\
        \midrule
        AT-DPT          & \colorMwA{115.5 $\pm$ 13.0}{114.0 $\pm$ 17.5}{111.1 $\pm$ 15.8}{114.9 $\pm$ 20.2} \\
        DPT             & \colorMwA{84.6 $\pm$ 13.8}{90.0 $\pm$ 14.8}{103.0 $\pm$ 12.5}{110.0 $\pm$ 14.7} \\
        PPO$^\S$        & \colorMwA{105.1 $\pm$ 8.3}{109.9 $\pm$ 9.0}{100.6 $\pm$ 5.3}{123.5 $\pm$ 8.1} \\
        \midrule
        \multicolumn{5}{c}{$\varepsilon = 0.4$} \\
        \midrule
        AT-DPT          & \colorMwA{104.8 $\pm$ 16.0}{116.8 $\pm$ 18.8}{108.6 $\pm$ 15.1}{112.7 $\pm$ 23.9} \\
        DPT             & \colorMwA{81.2 $\pm$ 12.2}{70.2 $\pm$ 15.0}{102.7 $\pm$ 13.1}{110.0 $\pm$ 14.7} \\
        PPO$^\S$        & \colorMwA{83.5 $\pm$ 7.4}{83.8 $\pm$ 7.2}{92.9 $\pm$ 7.3}{123.5 $\pm$ 8.1} \\
        \bottomrule
    \end{tabular}
    \label{tab:miniworld_full}
\end{table*}
}
\setlength{\tabcolsep}{6pt}

\subsection{Training curves}

\begin{figure}[!h]
    \centering
    \includegraphics[width=0.8\linewidth]{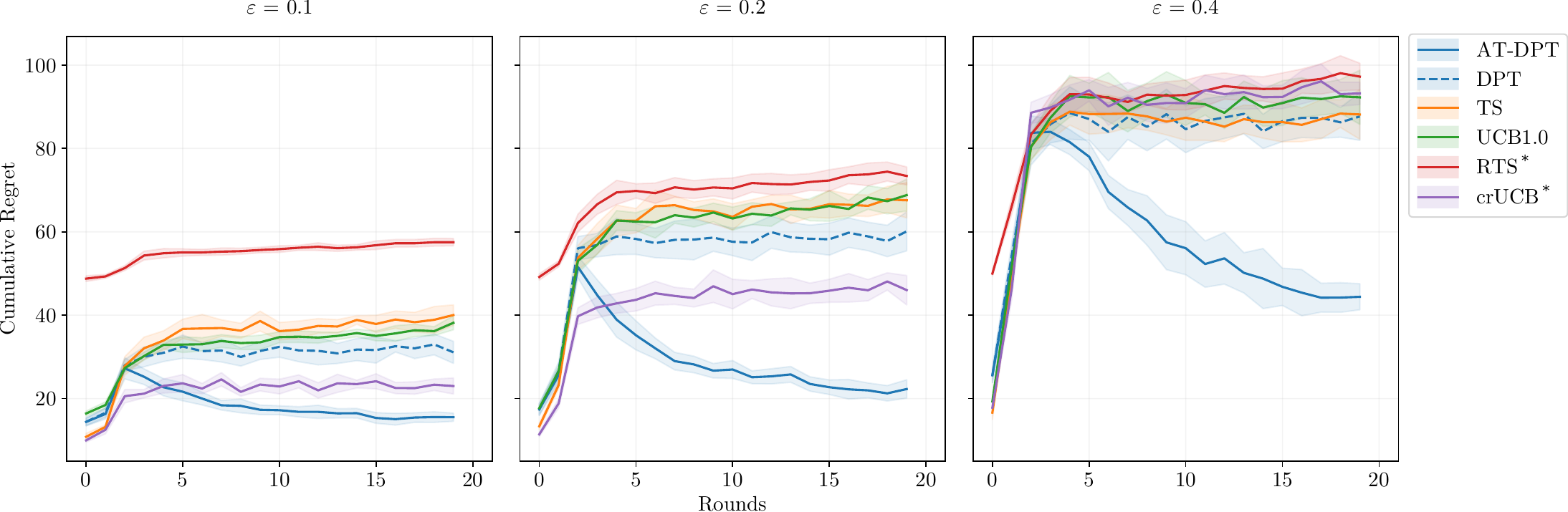}
    \caption{Adversarial training curves for training the attacker in the bandit setting, for different values of $\varepsilon$. Note, that in the case of AT-DPT it is trained along with the attackers. $^*$ We use tuned versions of RTS and crUCB, see \Cref{app:rts,app:crucb} for more details.}
    \label{fig:training_curves_bandit}
\end{figure}

\begin{figure}[!h]
    \centering
    \includegraphics[width=0.8\linewidth]{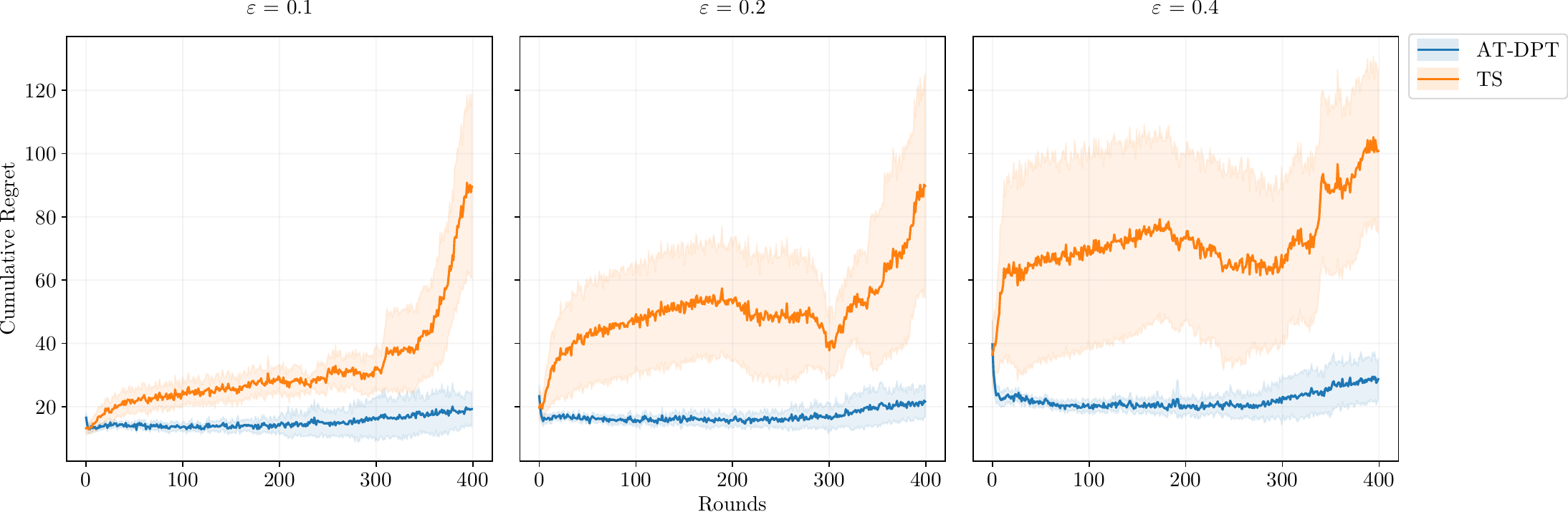}
    \caption{Adversarial training curves for training the adaptive attacker in the bandit setting, for different values of $\varepsilon$. Note, that in the case of AT-DPT it is trained along with the attackers.}
    \label{fig:training_curves_bandit_adaptive}
\end{figure}

\begin{figure}[!ht]
    \centering
    \includegraphics[width=\linewidth]{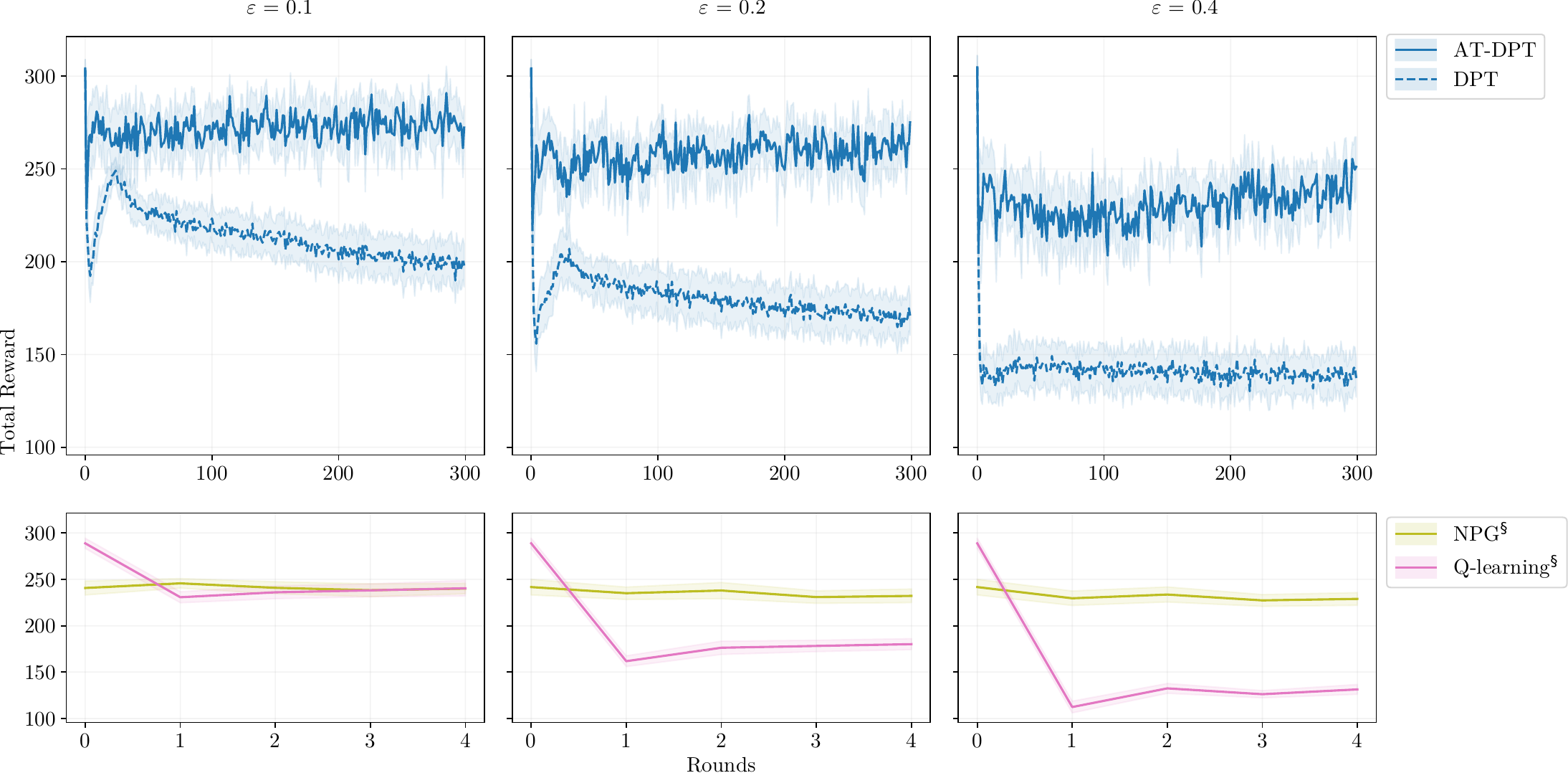}
    \caption{Adversarial training curves for training the attacker in the Darkroom2 environment, for different values of $\varepsilon$. Note, that in the case of AT-DPT it is trained along with the attackers. $^\S$ NPG and Q-learning require multiple episodes of online learning to converge to a stable policy; we run them for 100 episodes before evaluating their performance.}
    \label{fig:training_curves_darkroom}
\end{figure}

\begin{figure}[!htb]
    \centering
    \includegraphics[width=\linewidth]{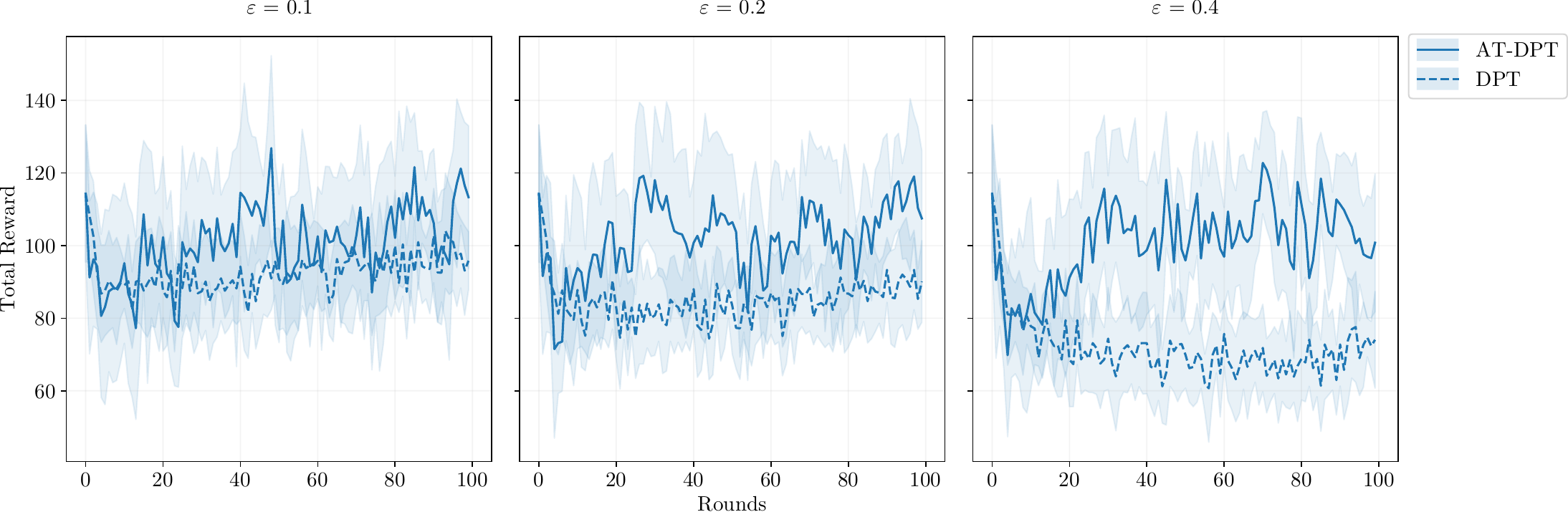}
    \caption{Adversarial training curves for training the attacker in the Miniworld environment, for different values of $\varepsilon$. Note, that in the case of AT-DPT it is trained along with the attackers.}
    \label{fig:training_curves_miniworld}
\end{figure}

\clearpage
\section{Further details}

\subsection{Compute resources}
The experiments were run on a compute cluster with NVIDIA A100 80GB PCIe and NVIDIA H100 94GB NVL GPUs. \\
Approximate GPU machine runtime of experiments, per run: \\[-1em]
\begin{tabular}{p{0.45\textwidth}p{0.45\textwidth}}
    \begin{itemize}
        \item Bandit Environment:
        \begin{itemize}
            \item Pretraining -- 3.4 h
            \item Adversarial training -- 0.4 h
            \item Evaluation -- 0.6 h
        \end{itemize}
        \item Bandit Environment, Adaptive Attacker:
        \begin{itemize}
            \item Pretraining -- 3.4 h (same as bandit env.)
            \item Adversarial training -- 0.6 h
            \item Evaluation -- 0.2 h
        \end{itemize}
    \end{itemize}
    & 
    \begin{itemize}
        % \item Linear Bandit Environment:
        % \begin{itemize}
        %     \item Pretraining -- 
        %     \item Adversarial training -- 
        %     \item Evaluation -- 
        % \end{itemize}
        \item Darkroom2 Environment:
        \begin{itemize}
            \item Pretraining -- 1.4 h
            \item Adversarial training -- 0.6 h
            \item Evaluation -- 3.4 h$^\S$
        \end{itemize}
        \item Miniworld Environment:
        \begin{itemize}
            \item Pretraining -- 13.1 h
            \item Adversarial training -- 2.7 h
            \item Evaluation -- 0.9 h
        \end{itemize}
    \end{itemize}
\end{tabular}\\[-1em]
$^\S$ NPG and Q-learning required multiple episodes of online learning before converging to a stable policy, therefore leading to an increased evaluation run time.

\subsection{Interpretation of attack in Darkroom2}
\begin{figure}[!htb]
    \centering
    \includegraphics[width=0.8\linewidth]{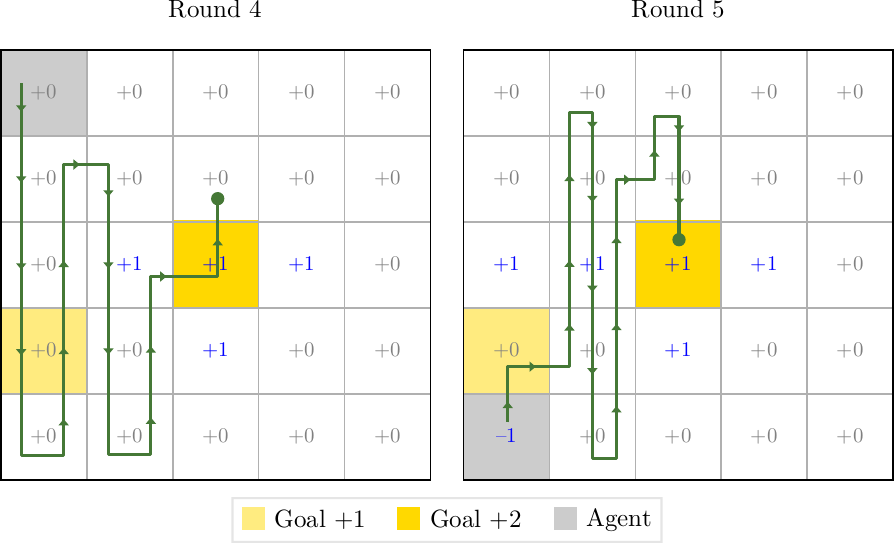}
    \caption{
        An illustration of the Darkroom2 environment with an attacker's poisoning strategy during a sample training run.
        Gray and blue numbers $-1$, $+0$, and $+1$ indicate the attacker's current poisoning strategy.
        Green path denotes trajectory taken by the agent for that round; green circle indicates the state where the agent chose to stop and exploit the current reward by choosing the stay action.
    }
    \label{fig:attack-interpret}
\end{figure}

We present an illustration of an example environment and attacker's strategy in \Cref{fig:attack-interpret}, taken from the middle of a sample AT-DPT adversarial training run.
The attacker's strategy observed in the illustration shows the attack is not arbitrary -- it is focusing on states nearby the goal.
We can see an attack of $+1$ on a goal which gives a reward of $2$ -- this would change the observed reward into $3$.
A reward value of $3$ was not seen during pretraining DPT, and in this round, upon encountering this it provokes undesirable behavior (\textit{stay} at a low-reward state), causing a low episode reward.
During the next round of training we find that the DPT has learned to recover from this mistake, and given the same attacker's strategy for that state successfully ignores this attack.

\clearpage
\subsection{AT-DPT test phase}
\begin{algorithm}[htb]
\caption{AT-DPT test phase}
\label{alg:atdpt-deployment}
\begin{algorithmic}[1]
    \STATE \textbf{input:} victim $\pi_\theta$ -- AT-DPT with params $\theta$
    \STATE \textbf{input:} attacker $\pidagger_\phi$ with params $\phi$, budget $B$, fraction of steps poisoned $\varepsilon$
    \STATE Sample $M$ tasks $\{ \Mcal_i \sim \Tcal \}_{i=1}^m$
    %\color{green!50!black} \quad $\triangleright$ Differing from the tasks in adversarial training \color{black}
    \FOR{all $\Mcal_i$ simultaneously}
        \STATE $s_0 \sim \rho_{\Mcal_i}$
        \STATE $D^\dagger \leftarrow \{\}$
        \FOR{$h = 0, \dots, H - 1$}
            \STATE select action $a_h \sim \pi_{\theta_{n}} (\,\cdot \mid D^\dagger, s_h)$
            \STATE $\rtilde_h = \begin{cases}
                \rdagger_h \sim \pi^\dagger_\phi(\,\cdot \mid s_h, a_h, \rbar_h) & \text{with probability } \varepsilon \\
                \rbar_h \sim R(\,\cdot \mid s_h, a_h) & \text{otherwise}
            \end{cases}$
            \STATE $s_{h+1} \sim T(\,\cdot \mid s_h, a_h)$
            \STATE append $(s_h, a_h, \rtilde_h, s_{h+1})$ to $D^\dagger$
        \ENDFOR
    \ENDFOR
\end{algorithmic}
\end{algorithm}

\end{document}